\documentclass[11pt]{article}

\usepackage[final]{acl}

\usepackage{times}
\usepackage{latexsym}
\usepackage[T1]{fontenc}
\usepackage[utf8]{inputenc}
\usepackage{microtype}
\usepackage{inconsolata}
\usepackage{graphicx}
\usepackage{booktabs}
\usepackage{amsmath}
\usepackage{amssymb}
\usepackage{cleveref}
\usepackage{multirow}
\usepackage{enumitem}
\usepackage{comment}
\usepackage{bm}
\usepackage{subcaption}
\usepackage{booktabs}
\usepackage{makecell}

\Crefname{equation}{Eq.}{Eqs.}

\title{Modeling the Arrival Time of Information-Bearing Events in Natural Language}
\author{%
    Leyi Lao\footnotemark[1] \:
    Yulin Ou\footnotemark[1]\:
    Yang Xu\footnotemark[2]\:
    Hao An\\
    Computational Linguistics and Consciousness Sciences Lab,\\ Southern University of Science and Technology, Shenzhen, China
}

\begin{document}
\maketitle

\begin{abstract}
Understanding how information is distributed over time is fundamental to modeling natural language. While existing empirical studies primarily rely on post-hoc analysis of information-theoretic metrics, such as surprisal, little is known about the generation processes under the surface.
% While prior work has focused on deterministic patterns such as uniform information density, the underlying temporal dynamics that govern when new information emerge during production remain underexplored.
% We model language generation as a stochastic process in which the discrete ``events'' that carry key information in the context emerge over time, with a learnable emergence strength. Events are defined with two different methods: simple rule-based \emph{new words} and probabilistic-based \emph{high-surprisal tokens}. Based on the between-event intervals, neural TPPs are learned to predict the timing of future events from the history, i.e., when the next new-word or high-surprisal-token is likely to occur. The learning is also augmented with contextual representations from LLMs.
We propose a novel approach for addressing this gap using the tool of neural temporal point processes (TPPs) by modeling the key information-bearing events in text sequences with learnable event intensity functions. Our model takes the history of event intervals and the contextual semantic representations from LLMs as input, and predicts the future arrival times for two types of events: \emph{new-word} and \emph{high-surprisal-token}. 
% Our main findings are: The emergence strength of information events can be learned by neural TPPs, which reveals a decaying pattern shared among human and LLM generations. However, this common pattern is only visible when events are defined by new words but not by high surprisal words. More interestingly, the raw values of emergence strength (measured by the intensity function in TPPs) provides an effective means to distinguish human and LLM generations: human strength is lower and decays faster, which roughly translates to -- LLMs have a higher tendency to introduce new words into the context. 
The experimental results reveal an interesting distinction in event intensities: LLMs tend to produce new words at significantly higher rates than humans, while high-surprisal tokens do not exhibit the difference. 
A decaying pattern of intensity over time is found in new-word events for both humans and LLMs, but not in high-surprisal-token events. 
Our findings revisit and extend previous studies that use simpler methods; provide supplementary evidence for theories about constrained context representation inherent in human language production; and find a new angle to view human and LLM generation differences. 
\end{abstract}

\section{Introduction}
\label{sec:introduction}
\footnotetext[1]{Equal contribution}
\footnotetext[2]{Corresponding author, email: xuyang@sustech.edu.cn}
Human language production involves not only choosing what comes next, but also when particular kinds of information enter the unfolding discourse.
For example, poems and public speeches have distinct rates of introducing new words \citep{badalamenti1994poisson,badalamenti1994lawful}; listeners in a conversation create smooth turn-takings by predicting the ending time of a speaker's current turn \citep{LEVINSON20166}.
In these cases, both new words and end-of-turn can be viewed as ``information-bearing'' events, whose timing is part of the distinct properties or functions of the language production process.
However, existing language modeling (LM) techniques can only characterize the distribution of these events in probability space, which is no different from other regular lexical tokens. Fewer technical lenses are available for specifically studying the \emph{temporal} arrival process of events.

The earliest relevant works we found in literature are from \citep{badalamenti1994poisson,badalamenti1994lawful,strand1997systems,badalamenti2001speech}, who treated the first occurrences of words in poems and speech transcripts as events, and found that their waiting times can be modeled with a slowly evolving Poisson process. Their findings established that information in language can indeed be studied through an evolving event rate.
However, from today's perspective, their methodology should be revisited: first, natural language data are heavy-tailed with ``bursts and lulls'' \citep{altmann2009beyond} and contain self-exciting dynamics such as long-range correlations \citep{altmann2012origin}, which motivates the need for more sophisticated modeling, instead of models in fixed forms (such as Poisson); second, the criteria for defining an ``event'' should be examined more broadly, extending from simple novelty-based to information metric-based ones. % Natural language data exhibit heavy-tailed distributions with ``bursts and lulls'' \citep{altmann2009beyond} and self-exciting dynamics inducing long-range correlations \citep{altmann2012origin}, which motivates the need for using more sophisticated models to study the intrinsic temporal signatures of language.

Modern neural temporal point processes (TPPs) combined with LLMs offer a pathway for the above-posited revisit.
Neural TPPs come with an abundance of effective architectures that have been proved mature solutions for learning the underlying stochastic processes behind sequence data of events, which usually involves a sequence modeling component (RNN or Transformer) plus some fully-connected layers as an event predictor \citep{du2016recurrent,omi2019fully,zuo2020transformer}.
Into the age of LLMs, there have been some recent endeavors on utilizing the contextual representation capabilities of LLMs to enhance the performance in predicting future events (such as re-tweets, online user behaviors) \citep{liu2024tpp,kong2025language}. % the \emph{intensity function} that quantifies the time-dependent event rate in time series
However, the current paradigm still needs be tailored to better align with the research interests of computational linguists -- why not go beyond the externally defined real-world events, and let the target of prediction be \emph{language} itself? 
% It is the goal of this study to explore the potentials of this new direction. 
% This paradigm has not been introduced to the field of natural language process, especially regarding the arrival times of information-bearing events during language generation.

This paper is an empirical attempt to bridge this gap.
We operationalize natural language as a temporal point process of discrete information-bearing events arriving along the time axis, where time is measured by word/token positions. We extract two types of events from the sequence of natural language: simple rule-based \emph{new-word}; and probability-based \emph{high-surprisal-token}. We implement the TPP model with an RNN architecture that learns the conditional intensity function of events from an encoded past history, and makes predictions on future event timing.
% that governs when the next event is likely to occur given the history of past events

% \citep{SOULE197457,DBLP:conf/naacl/Hale01,levy2008expectation,DBLP:journals/llc/Hale16}.
% xy todo: Move the citation of surprisal theory to Related Work. 
% Event arrival is conditioned on a mixed representation of inter-event intervals, absolute time step, and the events' contextual embeddings extracted from pre-trained LLMs, encoded with both recurrent and self-attentive history encoders. % todo: more specifically, RNN as encoded - TBD?

% xy TODO: Currently AI-summarized. Need polishing. 
Our main findings are as follows:
\begin{itemize}[leftmargin=10pt]
    \item \textbf{Event arrival time can be predicted accurately.}   
    Our neural TPP models are trained against text data from both spoken and written modalities, and the arrival time of a broad set of information-bearing events in human language can be accurately predicted, with better performance than conventional TPPs and simple neural baselines. 
    % we draw empirical conclusions extending the previous findings to a broader scope: 
    % generalizing the classical stochastic process perspective to multiple event definitions, including high-surprisal tokens.
    
    \item \textbf{LLMs can augment TPP modeling.} We find that the neural TPPs' modeling performance can be augmented by incorporating an LLM's hidden state encoding of history tokens. 
    It confirms that the internal representations of LLMs contain useful information for predicting the \emph{timing} of future tokens. 
    % The rich semantic contextual information from LLMs improves the event prediction performance than conventional non-LLM TPPs. 
    
    \item \textbf{Revealing human v.s. LLM differences in information dynamics.} 
    % By training neural TPPs over event sequences extracted from human-written and LLM-generated texts, 
    We find significant differences between human-written and LLM-generated texts in the \emph{intensity} of information arrivals. New-word events' arrival rates generally decay over time, but the rate of LLMs is consistently higher than  humans, which can be explained by the cognitive constraint unique in human language production; whereas the rates of high-surprisal-token events are close to flat and are less different. 
    % yield substantially flatter trajectories.
    % It provide a distinct and interpretable signal for understanding the differences in the generation processes behind.  
    % most information-bearing events exhibit a robust decaying intensity trend across both human and LLM texts.

    % \item \textbf{Potential new directions for training LLMs}
    % \item \textbf{Downstream evidence for the informativeness of intensity trajectories.} Features derived from the learned intensity functions---particularly temporal characteristics such as volatility and the number of peaks---support accurate and interpretable human--LLM classification across genres and generation models.
    
\end{itemize}
% Together, these results suggest that modeling the stochastic timing of information events offers a novel and interpretable lens for understanding the limits of current language modeling and natural language generation technology. 
% understanding and comparing human and machine language generation.

% - xy: when discussing human-llm distinctions, some refs may be needed.
% Temporal patterns which is recently found useful in large language model reasoning traces \citep{Gwak}.
% - The way human speakers plan information (such as new words, key information) are bounded by certain cognitive mechanisms, which can be characterized by mathematic models that predicts the timing of future ``event'', such as temporal point process.
% - It is not clear whether LLM-generation has the same underlying mechanism. It is question worth looking at because LLMs may or may not be the same as human speakers in terms of the inclinations/capabilities in distributing information temporally.
% - For the purpose of theory-development, we need more tools to describe the `dynamics` of information in natural lanugage, other than current spatial-time analysis tools. (UID, HR, Fourier etc.)

\section{Temporal Point Processes Formulated}
A temporal point process is a stochastic process that models the occurrence of discrete events over time. Formally, within a time window $[0,T]$ ($T\in\mathbb{R}^+$), the occurrence times of $N$ events are denoted by $\mathcal{H}=\{t_i\}^{N}_{i=1}$ ($i,N\in \mathbb{Z}^+$). Event $i$ occurs at time $t_i$, and so on.
At moment $t\in[0,T]$, let $\mathcal{H}_t=\{t_i|t_i<t\}$ be the history of events up to, but not including $t$, then the total number of events up to $t$ is denoted by a time-dependent random variable $N(t):=|\mathcal{H}_t|$, which is also known as a counting process.

The goal is to infer when the next event will occur, that is, to get an informed guess of the ($i+1$)-th event $t_{i+1}$ based on $\mathcal{H}_t$. This is done by learning a \emph{conditional intensity function} $\lambda(t|\mathcal{H}_t)$, which is the instantaneous rate that $t_{i+1}$ has \emph{not} happened before $t$, but will happen within a small interval right after, $[t,t+\Delta)$:
\begin{equation}\label{eq:intensity_func}
    \lambda(t|\mathcal{H}_t) = \lim_{\Delta \rightarrow 0}\frac{\mathbb{P}(t_{i+1}\in [t,t+\Delta)|\mathcal{H}_t)}{\Delta}.
\end{equation}

If $\lambda(t|\mathcal{H}_t)$ can be learned, then we can estimate the probability density function of the next event's occurrence time $t_{i+1}$:
\begin{equation}\label{eq:next_event_PDF}
\begin{split}
    p(t_{i+1} \mid \mathcal{H}_{t_{i+1}}) &= \\
    \lambda(t_{i+1} \mid \mathcal{H}_{t_{i+1}}) &\exp\left(-\int_{t_i}^{t_{i+1}} \lambda(t \mid \mathcal{H}_t)\, dt\right).
\end{split}
\end{equation}
which can be understood intuitively: the likelihood of seeing an event at $t_{i+1}$ is proportional to the instantaneous intensity (the $\lambda(\cdot)$ term on the right) multiplied by the likelihood of ``no events will occur during $[t_i,t_{i+1})$'' (the exponential and integration term). \Cref{eq:next_event_PDF} is a classical result in survival analysis, and it serves as the basis for constructing the learning objective of TPPs. We provide a brief derivation in \Cref{sec:A1-1_proof_PDF}. 

Conventional TPPs and neural TPPs mainly differ in how exactly the conditional intensity function $\lambda(t|\mathcal{H}_t)$ is parameterized, which will be introduced in the following two sections. 

% Next, we specify the definitions of ``event'' in the context of token-by-token natural language generation.

\subsection{Conventional TPP}
\label{sec:conventional_tpp}
Conventional TPPs rely on fixed-form parametric estimations $\lambda(t|\mathcal{H}_t)$. A common example is the stationary Poisson process, where $\lambda(t) = \lambda_0$ is constant over time, implying that all events occur independently at a fixed rate.
However, the stationary Poisson process is insufficient to capture the \emph{non-stationary} temporal patterns in real-world sequential data, such as natural language.
A solution is to use a non-homogeneous Poisson process with an intensity function $\lambda(t)$ that varies over time, such as the exponential decay proposed in \citep{badalamenti1994poisson}. 

Following this practice, we fit the cumulative rate function $\lambda_C(t)$ of information-bearing events (defined in \Cref{sec:def_events}) against the histogram of waiting times between events up to time $t$. $\lambda_C(t)$ is parameterized as a sum of two exponential terms plus a constant term:
\begin{equation}
\label{eq:cumulative_rate_function}
\lambda_C(t) = \lambda_1 e^{-\lambda_2 t} + \lambda_3 e^{-\lambda_4 t} + \lambda_5,
\end{equation}
where $\lambda_i \in\mathbb{R}^+$. Then the event intensity function $\lambda(t)$ can be estimated by taking the derivative: 
\begin{equation}
\begin{split}
\lambda(t) &= \frac{d \left(t\cdot\lambda_C(t)\right)}{dt} \\
& = \lambda_1 (1 - \lambda_2 t) e^{-\lambda_2 t} \\
& +\lambda_3 (1 - \lambda_4 t) e^{-\lambda_4 t} + \lambda_5.
\end{split}
\label{eq:convTPP_lambda_t}
\end{equation}
% $\lambda(t)=\frac{d}{d t}\left[t \cdot \lambda_{C}(t)\right]$.
% The true instantaneous rate $\lambda(t)$ is then obtained by differentiating the product of elapsed time t and the cumulative rate $\lambda_C(t)$, following the relation $\lambda(t)=\frac{d}{d t}\left[t \cdot \lambda_{C}(t)\right]$ \cite{badalamenti1994poisson}. Applying the product rule yields:

% KS test for \lambda_C, least squares when fitting, appling per-sequence, should these details be included?

This is a closed-form intensity function whose parameters have psychological meanings: $e^{-\lambda_2t}$ and $e^{-\lambda_4t}$ indicate two decaying rates; their strengths are regulated by the linear coefficients with $\lambda_1$ and $\lambda_3$; the constant $\lambda_5$ indicates an asymptotic rate near the end of text. The main reason of \citet{badalamenti1994poisson} setting two decaying rates rather than one is it better fit their data, and thus support their theories about the \emph{three phases} of human speech. Readers of interest can refer to the original study for more information. 
% \footnote{The brief idea is to lSet the second-order derivative $\lambda'(t)=0$ and solve for the zeros. Properly fit parameters will results in 1 or 3 zeros of $\lambda'(t)$, which characterizes the $\lambda(t)$ curve with three phases: \emph{gusto}$\rightarrow$\emph{fatigue}$\rightarrow$\emph{refresh}. }

\subsection{Neural TPP}
Neural TPPs allow for much more flexibility than conventional TPP in capturing the complex time-dependencies of the conditional intensity function $\lambda(t|\mathcal{H}_t)$. Their core advantage is in modeling the historical information with a neural network. 
We take the recurrent neural network (RNN)-based TPPs proposed in \citet{omi2019fully} for our main implementation, 
in which the event history $\mathcal{H}_t$ is encoded with an RNN hidden state  $\bm{h}_i$ (for $i$-th event). 
The conditional intensity is now formulated as a function of the hidden state $\bm{h}_i$ and the elapsed time since the last event $\tau = t - t_i$, as follows:
\begin{equation}
\lambda(t|\mathcal{H}_t) \triangleq \phi(\tau|\bm{h}_i),
\label{eq:hazard_func}
\end{equation}
where $\phi(\cdot)$ is a non-negative function referred to as the \textbf{hazard function}, intuitively meaning the ``risk'' of seeing an event at moment $t$. 

% Traditional RNN-TPP models often impose specific forms on $\phi$, such as an exponential form $\phi(\tau|\bm{h}_i) = \exp\left( w^\tau \tau + \mathbf{v}^\top \bm{h}_i + b \right)$ or a constant form (corresponding to exponentially distributed inter-event times). However, these assumptions restrict the model's ability to adapt to complex temporal patterns.

% xy TODO: (Transformer-based model is also implemented, see Appendix...)
% \citep{du2016recurrent,omi2019fully}.
% Unlike conventional TPPs, which impose strong assumptions on the forms of conditional intensity function, 

As pointed out in \citet{omi2019fully}, the loss function of learning a TPP requires an integral term of the hazard function (it can be seen by substituting \Cref{eq:hazard_func} into \Cref{eq:next_event_PDF}), which limits the generalizability of the model. The solution they have proposed is to learn the \textbf{cumulative hazard function} instead:
\begin{equation}
\Phi(\tau|\bm{h}_i) = \int_0^\tau \phi(s|\bm{h}_i) \, ds,
\label{eq:cumulative_hazard_def}
\end{equation}
% which intuitively means the cumulative amount of probability of seeing an event up to elapsed time $\tau$ since the last event at $t_i$. 
which intuitively means the accumulated risk or the expected number of events up to elapsed time $\tau$ since the last event at $t_i$. 

In the simplest form of neural TPPs, the RNN receives the previous hidden state $\bm{h}_{i-1}$ to compute the current state $\bm{h}_i$, which, together with the elapsed time $\tau$, connect to a positive-valued feedforward neural network that outputs a predicted value of $\Phi(\tau|\bm{h}_i)$. This prediction is then used against the ground truth event times $\{t_i\}^{N}_{i=1}$ to compute the final loss $\mathcal{L}_{\text{NTPP}}$. 
$\mathcal{L}_{\text{NTPP}}$ is defined by maximizing the log-likelihood of \Cref{eq:next_event_PDF}, which is different from the cross-entropy or mean-squared-error loss commonly seen in language modeling (Formulation given in \Cref{eq:AppA-2_NTPP_loss} in \Cref{sec:A1-2_NTPP_loss}). 
% Once a neural TPP is trained, we can easily compute the event intensity, i.e., hazard, via automatic differentiation: $\phi(\tau|\bm{h}_i) = \frac{\partial}{\partial \tau} \Phi(\tau|\bm{h}_i)$. 
% \begin{equation}
% \label{eq:phi_derivative}
% \phi(\tau|\bm{h}_i) = \frac{\partial}{\partial \tau} \Phi(\tau|\bm{h}_i)
% \end{equation} 

To sum up, a neural TPP model requires no prior assumptions on the form of the intensity function, and thus in theory can express complex temporal dynamics better than the conventional TPP. 
Additionally, we also experiment with transformer-based neural TPPs, detailed in \Cref{sec:tansformer-based}.
Building upon this foundation, we apply neural TPPs to natural language and integrate them with LLMs to model event intensity not only from the historical times encoded in $\bm{h}_i$, but also from rich contextual semantics.
% One of our goals is to integrate neural TPP with LLMs, so that the intensity of events can be modeled not only from historical times encoded in $\bm{h}_i$, but also from the rich event semantics encoded by LLMs. 
% Details will be introduced in the next section. 

% xy: commented out
% This approach requires no prior assumptions about the form of the intensity function, granting greater expressive ability. The integral in the likelihood function can be computed exactly via $\Phi$, avoiding numerical approximations.

\begin{figure*}
    \centering
    \includegraphics[width=.95\textwidth]{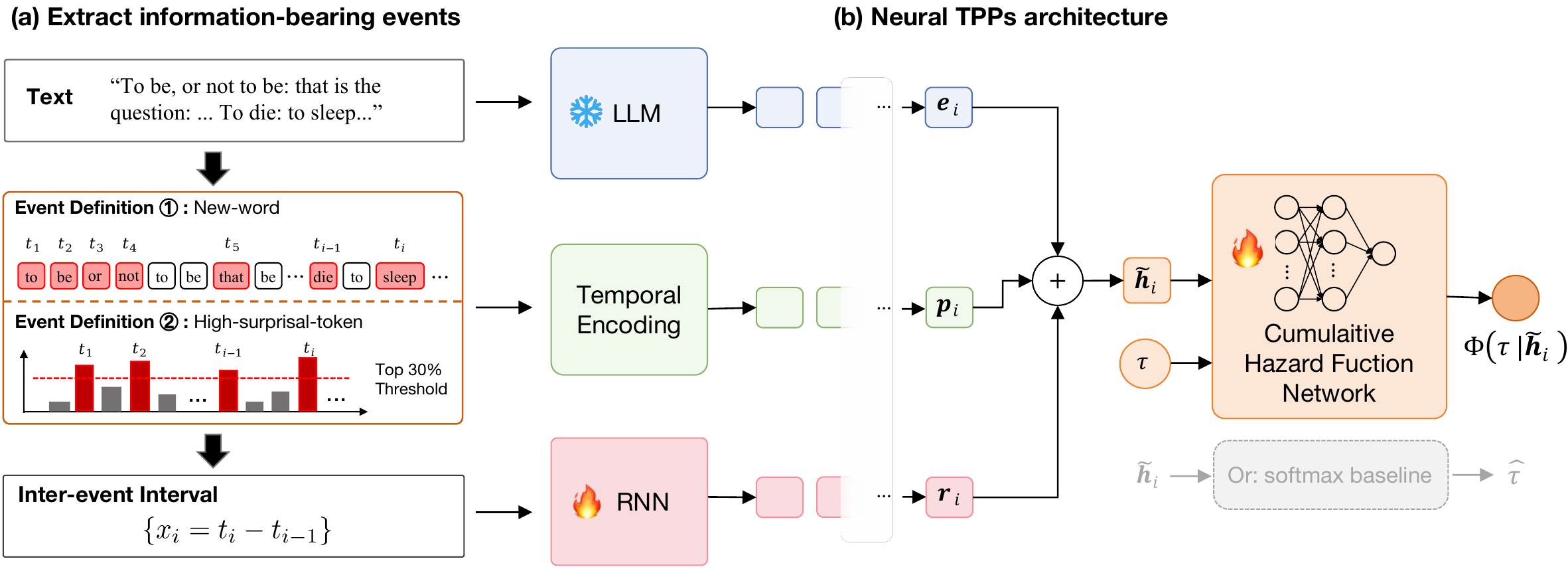}
    \caption{Overview of our RNN-based neural TPP framework. (a) Event extraction from text: new-word or high-surprisal-token. (b)Neural TPP architectures that encode inter-event intervals, semantic embeddings, and positional encodings to estimate the cumulative hazard function.}
    \label{fig:NTPP}
\end{figure*}

% xy TODO: perhaps move to intro, and add a simple example explaining the idea of how our expectation for future words is closely related to word meanings. 
% While general neural TPPs model events only in the temporal domain, linguistic events (such as the occurrence of new words or high-surprisal tokens) are deeply related to their semantic context.

\section{Neural TPP for Natural Language}
We adapt neural TPPs to natural language input in two steps. First, we define which positions in a text constitute information-bearing events, thereby converting the text into an event sequence. Then, we enrich the temporal history of events with contextual representations from an LLM and use the resulted mixture of both temporal and semantic information to predict the arrival time of the next event. Figure~\ref{fig:NTPP} provides an overview of the pipeline.

\subsection{Defining Information-Bearing Events}
\label{sec:def_events}
We begin by defining what can be counted as information-bearing events within the sequence of natural language.
For each event definition, the positions of the extracted events form an ordered sequence $\mathcal{H}=\{t_i\}_{i=1}^{N}$, where $t_i$ is the word or token position of the $i$-th event. % Is the definition of the time axis here necessary?
A natural choice is the previously investigated one, new words, i.e., the first occurrences of word types. Beyond that, we propose to use surprisal as a selection criterion, because it is a well-established metric that reflects the amount of information in linguistic units.

\begin{itemize}[leftmargin=10pt]
\item \textbf{New-word events.}
We define the occurrence of new words, i.e., words that have not appeared in the preceding context as events, which is the same setup as \citet{badalamenti1994poisson}. Nouns and verbs are lemmatized to the base forms.

\item \textbf{High-surprisal-token events.}
To capture the temporal patterns of information distribution, we define events based on token-level surprisal computed using LLaMA-3.1-8B~\citep{grattafiori2024llama3herdmodels}.
We aggregate the surprisal values of all tokens across all texts (including human and LLM-generated), and a token is marked as an event if its surprisal value falls into the top 30\% of this global distribution.
% To establish a fair and standardized threshold, 
\end{itemize}

By the two definitions, each text is converted into an event sequence that serves as the input data to the TPP model. The time axis is measured in word positions for new-word events and token positions (BPE tokenized) for high-surprisal-token events.

% \subsection{Complete Architecture}
\subsection{Context Augmented Architecture}
After extracting the events, we encode their temporal history and augment them with the semantic representations from LLMs. The motivation is straightforward: since the target events are occurrences of words (tokens), whose meanings should reflect an anticipated ``psychological'' distance of the yet-to-come future words.
Concretely, as illustrated in Figure \ref{fig:NTPP}, we condition the cumulative hazard function on an integrated historical representation, $\Phi(\tau|\tilde{\bm{h}}_i)$, where $\tilde{\bm{h}}_i$ consists of the three following components:

\begin{enumerate}[leftmargin=10pt]
    \item \textbf{Historical event intervals}: The sequence of intervals $x_i = t_i - t_{i-1}$ captures the temporal spacing between events. Practically, we set a sliding window of size $W$ and use the segment of most recent intervals $[x_{i-W+1}, \dots, x_i]$ as the inputs to an RNN. The RNN learns the temporal patterns among the intervals, and yields a hidden state $\bm{r}_i$ that encodes this knowledge: 
    $\text{RNN}(x_{i-W+1}\dots x_i):\rightarrow \bm{r}_i$.
    
    \item \textbf{Absolute time steps}: $t_i$ is the specific occurrence time of the $i$-th event within the word/token stream. In order to fuse with other sources of input, the integer time step $t_i$ is encoded to a continuous positional embedding (PE), which is similar to the Sinusoidal PE used in Transformer: $\text{PE}(t_i):\rightarrow \bm{p}_i$.

    % \item \textbf{Contextual semantic embedding}: We feed all the preceding text up to the $i$-th event (including the event word/token itself), $[w_1,w_2,\dots,w_{t_i}]$, to a pre-trained LLM. Note that the subscripts include all events and non-events tokens, that is, everything is taken as input. Then we extract the hidden state embedding $\bm{e}_i$ from the LLM's last layer at the last token position (roughly where $i$-event is at), as the contextual semantic representation for the event: $\text{LLM}(w_{1}\dots w_{t_i}): \rightarrow \bm{e}_i$.
    % 我有些担心会不会让人误以为(including the event word/token itself)使得input里包含了预测目标的语义信息，因为在下一节才明确我们要预测的是t_{t+1}。我让AI写了一版如下：
    \item \textbf{Contextual semantic embedding}: We feed all the preceding text up to and including the $i$-th event, $[w_1,w_2,\dots,w_{t_i}]$, to a pre-trained LLM. This prefix contains both event and non-event words/tokens. We then extract the last-layer hidden state at the final position as the contextual semantic representation of the event: $\text{LLM}(w_{1},\dots,w_{t_i}) \rightarrow \bm{e}_i$. Because only the prefix up to $t_i$ is used, the representation contains no information from future positions.
\end{enumerate}

Finally, we construct the integrated hidden state vector $\tilde{\bm{h}}_i$ by additively fusing the above three sources with corresponding learnable weights: 
% $\bm{h}_i = W_r\mathbf{r}_i +  + W_p\mathbf{p}_i$. 
\begin{equation}
   \tilde{\bm{h}}_i = W_r\bm{r}_i + W_p\bm{p}_i + W_e\bm{e}_i.
\end{equation}
The resulting vector is fed into the cumulative hazard function network to compute $\Phi(\tau|\tilde{\bm{h}}_i)$.

\subsection{Predicting Event Arrival Times}
Since the intensity functions' outputs only carry probabilistic meanings, we need to transform them into concrete predictions of arrival times for future events, upon which the prediction performance can be measured.
% The following ``trick'' is applied 
We apply conditional median prediction to map a real-valued intensity score to a predicted time. Assuming the previous event occurs at $t_i$, we aim for a predicted time $t^*_{i+1}$ at the median of the predictive distribution, such that the actual next event has a 50\% probability of occurring before it and a 50\% probability of occurring after it. 
For neural TPPs, it translates to this equation regarding the cumulative hazard: $\exp(-\Phi(\tau\mid\bm{h}_i))=1/2$, which is solved to $\Phi(t^*_{i+1}-t_i\mid \bm{h}_i)=\log(2)$.
For conventional TPPs, we simply replace $\Phi(\cdot)$ with $t\cdot\lambda_C(t)$, and obtain $t^*_{i+1}$ in the same manner. 
We then calculate the mean absolute error, $\text{MAE}=1/N\cdot\sum \lvert t^*_{i+1}-t_{i+1}\rvert$, to measure the predicting performance. 
This method of calculating and evaluating predictions from TPPs is a common practice adopted from \citet{omi2019fully}. 
% The exception is for our neural baselines, in which the arrival time is directly produced by taking the argmax of the softmax categorical output.

\section{Experimental Setup}

\subsection{Datasets and Models}
We use text data in two modalities: The \texttt{Speech} dataset that we collect from American Rhetoric\footnote{\url{www.americanrhetoric.com/speechbank.htm}}, an online high-quality resource of public speeches in English (200 samples, 310K words in total), and the well-established \texttt{Writing-Prompts} dataset \citep{writingprompts} for written texts (2000 samples, 2.32M words in total). % Speech 309,958; Writing-Prompts 2,337,315  
For preparing LLM-generated text data, we use Claude-Sonnet-4.5 \citep{anthropic2025claudesonnet45}, GPT-4.1 \citep{openai2024gpt4technicalreport}, and Qwen-Plus \citep{yang2025qwen3}\footnote{For all three proprietary models, the date of APIs being called is: 2025-10.}. Generation lengths are controlled to the same level as human dataset. 
% For the LLM-integrated TPPs, we employ two lightweight pre-trained models (LLaMA-3.2-1B and Qwen-3-0.6B), to extract the contextual semantic embeddings required by our neural TPP architectures.
For the LLM-integrated TPPs, we use LLaMA-3.1-8B to extract the contextual semantic embeddings for the main experiments.
\Cref{sec:appendix_dataset} describes details about data preprocessing. 
% and other embedding models we have experimented with.

\paragraph{Conventional TPPs training setup:}
We independently fit the cumulative rate function $\lambda_C(t)$ in eq. (\ref{eq:cumulative_rate_function}) to all texts within each dataset, which will yield a set of five parameters $\{\lambda_{1}\dots\lambda_{5}\}$ for each data sample. To rigorously assess the goodness-of-fit of this parametric model, following \citet{badalamenti1994poisson}, we evaluate the Residual Mean Square (RMS) of the regression and the correlation ($R$) between the fitted and observed data.

\paragraph{Neural TPPs training setup:}
We apply a 80\%-20\% train-test split to each dataset.
To achieve stable training on our discrete time steps (i.e., $\{t_i\}$ are integers), we augment the training data by injecting zero-mean Gaussian noise to each interval during training \citep{DBLP:journals/corr/TheisOB15, DBLP:conf/icml/HoCSDA19}.
Additionally, for computational efficiency, the extracted semantic embeddings $\bm{e}_i$ are projected onto a lower-dimensional space using PCA. %\citep{abdi2010principal}
The complete training configurations are provided in \Cref{sec:appendix_Implementation}. 
% (details in Appendix \ref{sec:PCA}).
% (details in Appendix \ref{sec:dequantization})
% \ref{sec:Configurations}.

% xy todo: move MAE def. to Section 5 Results part.
% Following the evaluation framework of \citet{omi2019fully}, we employ the Mean Negative Log-Likelihood (MNLL) and the Mean Absolute Error (MAE) as our primary evaluation metrics. To calculate the MAE, the timing of an upcoming event is predicted by finding the median of the prediction distribution via a root-finding method (detailed in Appendix \ref{sec:MAE}).

% \subsection{Neural Baseline}
%  To evaluate the advantages of neural TPPs, we construct a neural baseline for comparison. Instead of estimating a cumulative hazard function $\Phi$, this baseline simplifies the prediction of the next event time into a standard classification task,  conditioned on the identical historical hidden state $h_i$.
% Specifically, we remove the TPP integration framework and directly feed $h_i$ into a linear projection followed by a Softmax activation. Trained via standard cross-entropy loss, the model predicts the discrete inter-event interval $\tau$ as a categorical class. 
% This baseline allows us to rigorously assess the performance gains provided by the specific TPP mathematical formulation over naive sequence modeling.

\subsection{Classification Tasks}\label{sec:classification_tasks}
To further investigate the differences in temporal patterns between human and LLM-generated texts, we conduct a classification experiment using the intensity curves from neural TPP models as inputs. The procedure consists of three steps:

\paragraph{TPP as estimator:} Train a neural TPP model (with LLM embeddings) on a prompt set of 10,000 human-written stories. The obtained model is used as the intensity curve estimator. % LLaMA-3.2-1B embeddings as additional inputs.

\paragraph{Feature extraction:} Apply the trained estimator to a classification set that includes 200 speeches and 2,000 unique stories (each sample comes with 1 human version and 3 LLM versions), thus producing intensity curves for $4 \times 200 = 800$ speeches and $4 \times 2000 = 8000$ stories in total. Then we extract statistical features from these curves, including global statistics (e.g., mean, variance, skewness, kurtosis and percentiles), local statistics (mean and variance per time bin), and temporal characteristics (e.g., number of peaks, volatility and trend slope).

\paragraph{Classification \& analysis:} Finally, train a Logistic Regression classifier based on features extracted and perform a binary classification (Human vs. LLM). The goal is to quantify the difference in temporal patterns between the two sources, and use feature importance to analyze which aspects of the intensity curves are most discriminative.

\begin{comment}
\begin{enumerate}[leftmargin=10pt]
    \item \textbf{TPP training}: Train a neural TPP model (with LLM embeddings) on a prompt set of 10,000 human-written stories. The obtained model is used as the intensity curve estimator. % LLaMA-3.2-1B embeddings as additional inputs
    \item \textbf{Feature extraction}: Apply the trained estimator to a classification set that includes 200 speeches and 2,000 unique stories (each sample comes with 1 human version and 3 LLM versions), thus producing intensity curves for 4x200=800 speeches and 4x2000=8,000 stories in total. Then we extract statistical features from these curves, including global statistics (e.g., mean, variance, skewness, kurtosis and percentiles), local statistics (mean and variance per time bin), and temporal characteristics (e.g., number of peaks, volatility and trend slope).
    \item \textbf{Classification \& analysis}: Finally, train a Logistic Regression classifier based on features extracted and perform a binary classification (Human vs. LLM). The goal is to find whether temporal patterns differ significantly between the two sources, and use feature importance analysis to reveal which aspects of the intensty curves are most discriminative.
\end{enumerate}
\end{comment}

\begin{table*}[!htbp]
\centering
\caption{Comparison of predicting performance (MAE$\downarrow$) among conventional TPPs, neural baselines, and neural TPPs, with and without LLaMA-3.1-8B semantic embeddings. Conventional TPPs for high-surprisal-token are left empty due to the low rate of successful fit.}
\label{tab:TPP_mae_results}
\resizebox{\textwidth}{!}{ 
\begin{tabular}{ll|l|ccccc}
\toprule
\textbf{Dataset} & \textbf{Event Type} & \textbf{Text Source} & \textbf{\makecell{Conventional \\ TPP}} & \textbf{\makecell{Neural Baseline \\ w/o LLM}} & \textbf{\makecell{Neural TPP \\ w/o LLM}} & \textbf{\makecell{Neural Baseline \\ w/ LLM}} & \textbf{\makecell{Neural TPP \\ w/ LLM}} \\
\midrule

\multirow{8}{*}{\textbf{Speech}} & \multirow{4}{*}{\textbf{New-word}} 
& Human & 1.890 & 2.135 & 1.779 & 1.867 &  \textbf{1.726} \\
& & Claude-Sonnet-4.5 & 1.256 & 1.226 & 1.086 & 1.069  & \textbf{1.019} \\
& & GPT-4.1 & 1.299 & 1.286 & 1.118 & 1.043 &  \textbf{1.009} \\
& & Qwen-Plus & 1.162 & 1.154 & 1.000 &  0.878 &  \textbf{0.859} \\
\cmidrule(lr){2-8}

& \multirow{4}{*}{\textbf{\makecell{High-surprisal-\\token}}} 
& Human & - & 2.202 & 1.888 & 1.919 & \textbf{1.780} \\
& & Claude-Sonnet-4.5 & - & 2.201 & 1.785 & 1.764 & \textbf{1.624} \\
& & GPT-4.1 & - & 1.841 & 1.473 & 1.378 & \textbf{1.322} \\
& & Qwen-Plus & - & 2.718 & 2.121 & 2.175 & \textbf{2.003} \\
\midrule

\multirow{8}{*}{\textbf{\makecell{Writing \\ Prompt}}} & \multirow{4}{*}{\textbf{New-word}} 
& Human & 1.676 & 1.886 & 1.567 & 1.580 & \textbf{1.462} \\
& & Claude-Sonnet-4.5 & 1.496 & 1.607 & 1.361 & 1.376 & \textbf{1.265} \\
& & GPT-4.1 & 1.371 & 1.452 & 1.222 & 1.209 & \textbf{1.118} \\
& & Qwen-Plus & 1.362 & 1.393 & 1.201 & 1.134 & \textbf{1.075} \\
\cmidrule(lr){2-8}

& \multirow{4}{*}{\textbf{\makecell{High-surprisal-\\token}}} 
& Human & - & 2.094 & 1.743 & 1.738 & \textbf{1.569} \\
& & Claude-Sonnet-4.5 & - & 3.218 & 2.439 & 2.592 & \textbf{2.285} \\
& & GPT-4.1 & - & 1.995 & 1.604 & 1.521 & \textbf{1.413} \\
& & Qwen-Plus & - & 2.539 & 2.021 & 1.950 & \textbf{1.799} \\

\bottomrule
\end{tabular}
}
\end{table*}

\section{Results}

% xy todo:
\subsection{Revisiting the Previous Study on New-word Events} % Conventional + Neural TPPs.
\label{sec:temp_dynamics_of_new_word_events}
% 复现实验 lambda_t curve
% A general view of the difference between human and llm texts
% - 5-param lambdas
% - Outputs from hazard function
% The main point is to show that neural TPPs can produce an conditional intensity curve that is consistent with the previous study using conventional TPPs.
This section revisits \citet{badalamenti1994poisson}'s original findings and extends the analysis to both conventional and neural TPPs on larger datasets. 

% First, the estimated $\lambda_{i=1\dots5}$ parameters of the conventional TPP result in intensity functions $\lambda(t)$ (\Cref{eq:convTPP_lambda_t}) whose curves decrease over time (\Cref{fig:5-1_intensity}, left column), consistent with \citet{badalamenti1994poisson}'s results. 

\begin{figure}[h]
\centering
\begin{subfigure}[b]{0.5\textwidth}
    \centering
    \includegraphics[width=\linewidth]{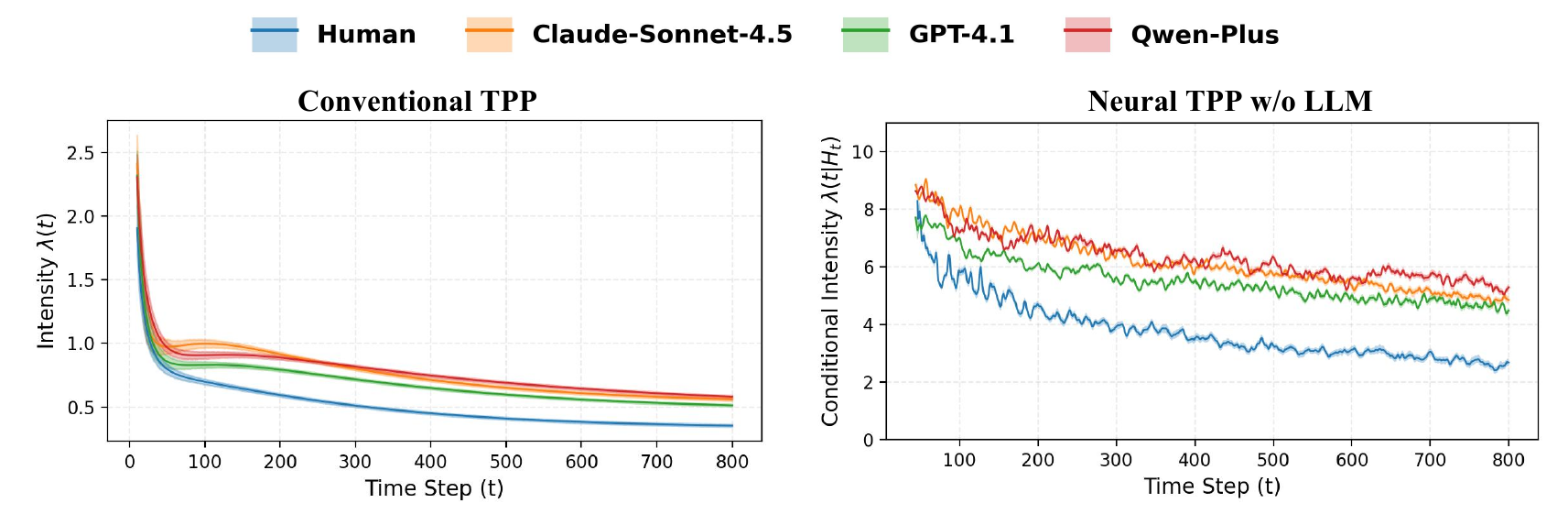}
    \caption{Speech dataset.}
    \label{fig:5-1_intensity_speech}
\end{subfigure}
\hfill
\begin{subfigure}{0.5\textwidth}
    \centering
    \includegraphics[width=\linewidth]{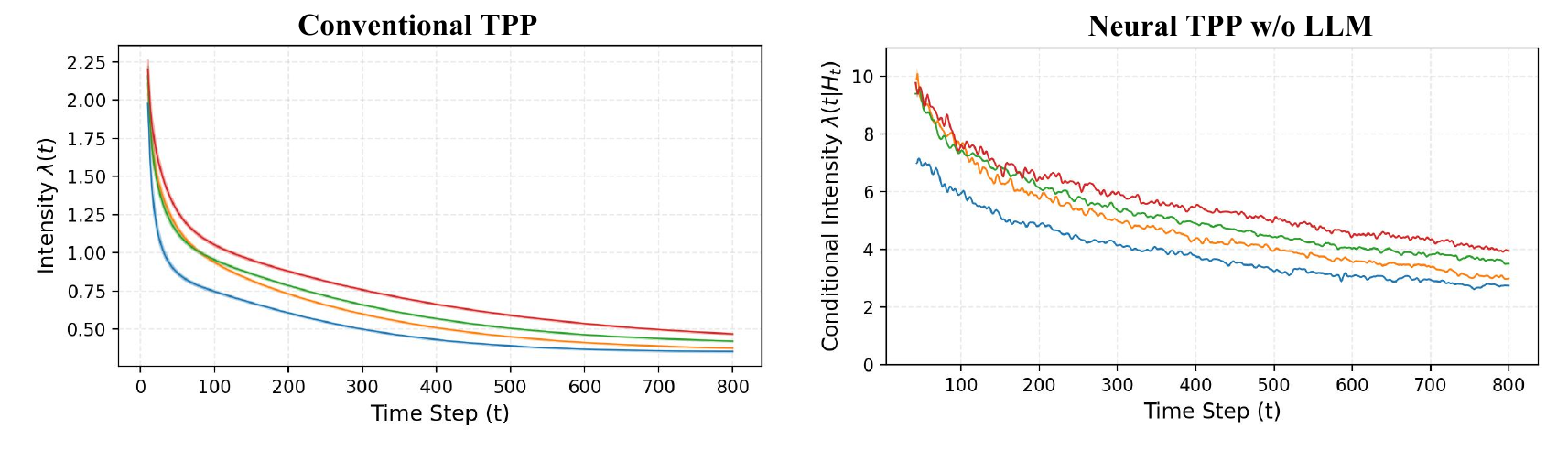}
    \caption{Writing-Prompts dataset.}
    \label{fig:5-1_intensity_writing}
\end{subfigure}
\caption{Estimated conditional intensity against time step for new-word events from conventional and neural TPPs. Both TPPs rely solely on temporal information without  semantic embedding input. Shadow areas are 95\% bootstrap confidence intervals.}
\label{fig:5-1_intensity}
\end{figure}

First, for the conventional TPP, we fit the $\lambda_{i=1\dots5}$ parameters for each text within every subset to obtain the sample-level intensity function $\lambda(t)$.
The goodness-of-fit is comparable to the original study ($\textit{RMS}\approx 0.0015$ and $R > 0.96$, see \Cref{sec:Conventional_TPPs_fitting}). 
We estimate the intensity scores for each time step over the input trajectory from the fitted parameters, and compute the average over data subset,  resulted in the final intensity curves (\Cref{fig:5-1_intensity}, left column). We find a general decay of $\lambda(t)$ over time, consistent with \citet{badalamenti1994poisson}'s results. 
% We also notice that the ``drop-increase-decrease'' U-shape pattern reported in \citep{badalamenti1994poisson} is only slightly present in the three model curves from Speech, but not in other groups. 

% xy todo: commented out for now.
% Specifically, every sample $j$ in a data subset, $\mathcal{D}_\text{Human}$ for instance, is used once to estimate the five parameters $\hat{\lambda}^{(j)}_{i=1\dots5}$. The per-sample intensity curve $\lambda^{(j)}(t)$ is then computed from $\hat{\lambda}^{(j)}_{i=1\dots5}$ via \Cref{eq:convTPP_lambda_t}, and the plotted curve is the trajectory average across the subset, $\bar{\lambda}(t)=1/|\mathcal{D}_\text{Human}|\sum_{j\in \mathcal{D}_\text{Human}}\lambda^{(j)}(t)$. 

Interestingly, the intensity curves from the human texts are consistently lower than those from the LLM texts, meaning that \textbf{LLMs tend to produce new words at higher rates than humans}, in both spoken and written texts. 
This difference is more pronounced in the conditional intensity curves returned from neural TPPs (\Cref{fig:5-1_intensity}, right column), further confirming the finding. 

Here, the conditional intensity curves from the neural TPPs are obtained by running inference with individual models trained on each subset, without the LLM embedding input $\bm{e}_i$, i.e., using only the inter-event intervals ($\bm{r}_i$) and absolute-time encodings ($\bm{p}_i$) as model inputs. 
This keeps the comparison on an equal basis. The plotted curves are average over per-sample hazard $\phi(\cdot|\bm{h}_i)$ at each time step across subset.
% , as both TPPs rely solely on temporal information without any semantic input. 
% To generate the neural TPP curves, at each new-word event $i$ in a text, whose occurrence time is $t_i$, the hazard $\phi(\tau|\bm{h}_i)$ (\Cref{eq:hazard_func}) is evaluated at integer $\tau$ up to the next event, and the output is assigned to the time step $t_i+\tau$. 
% To smooth minor local variations, the intensity at each step is averaged over a 3-point local window ($[t-1, t+1]$). 
% with 95\% bootstrap confidence intervals computed. 
 
The neural TPP curves exhibit steadier decay rates over the discourse, in contrast to the early quick drop in the conventional TPPs curves. 
Their curves are less smooth though. This is expected as RNN is supposed to be more sensitive to context than a deterministic function $\lambda(t)$. 
% For the complete intensity curves generated by neural TPPs integrated with LLM embeddings, which exhibit highly consistent temporal dynamics, please refer to \Cref{sec:D2-full_curves}.
We examine if this translates to better event predicting performance in \Cref{sec:NeuralTPP_adv}.

% Exhibit the temporal patterns described by \citet{badalamenti1994poisson}: an initial burst of new words followed by a general decline and a U-shaped recovery.
% xy todo: this can be put to discussion. 

% xy todo: for now; conventional tpp plots seem too flat?

\subsection{Advantages of Neural TPPs and LLM-Integration}\label{sec:NeuralTPP_adv}

Next, we thoroughly examine the advantages of neural TPPs over conventional TPPs and neural baselines, in terms of their performances in modeling event arrival times. 
Particularly, the neural baseline uses the identical historical hidden state $\bm{h}_i$ but bypasses the cumulative hazard $\Phi$ network, predicting the next-event interval $\tau$ directly via a standard softmax output.

% For accessing the advantages of integarting with LLM embeddings, we compare the MAE performances between three conditions: conventional TPPs, raw neural TPPs, and neural TPPs with LLM embeddings. 
To assess the advantages of neural TPPs and the integration of LLM embeddings, we compare the MAE performances across five conditions, as shown in \Cref{tab:TPP_mae_results}. 
We can see that the ``Neural TPP w/ LLM'' condition consistently achieves the lowest MAE, confirming that integrating LLM representations can significantly enhance the prediction of event arrival times, for both event types and datasets.
Also, despite sharing the same RNN encoder, neural baselines yield a higher MAE (lower accuracy) than neural TPPs, emphasizing the necessity of having an explicit cumulative hazard network in modeling language processes.

\begin{figure}
\centering
\begin{subfigure}[b]{0.5\textwidth}
    \centering
    \includegraphics[width=\linewidth]{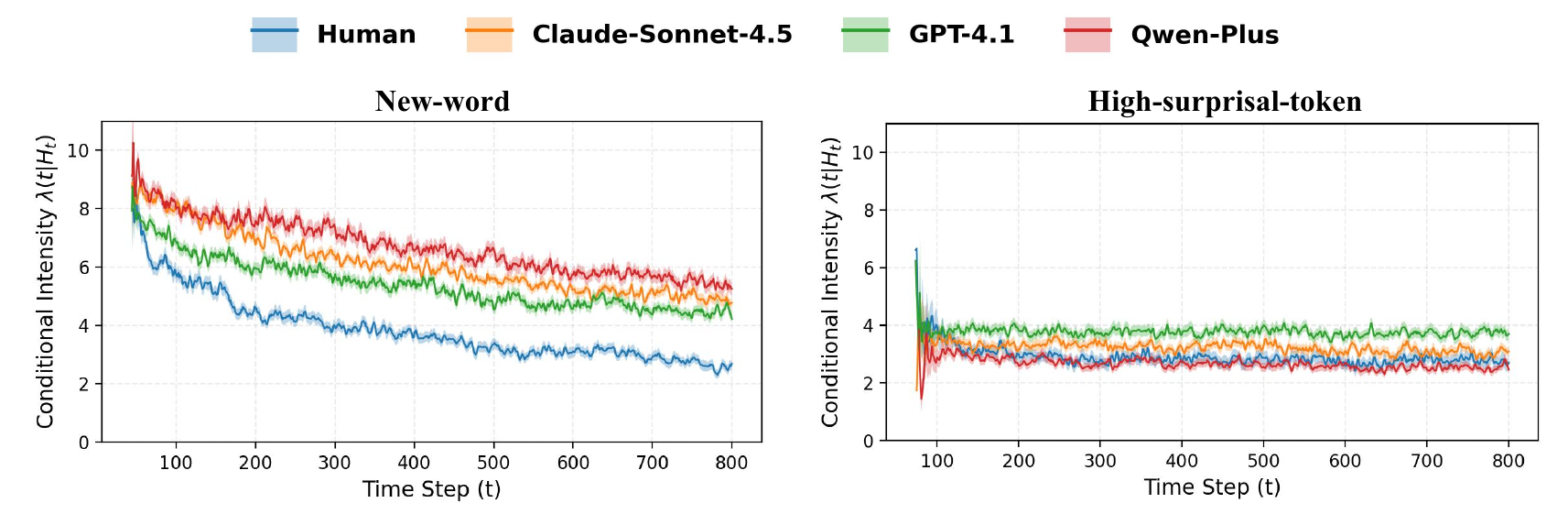}
    \caption{Speech dataset.}
    \label{fig:5-3_speech_words_vs_entropy}
\end{subfigure}
\hfill
\begin{subfigure}{0.5\textwidth}
    \centering
    \includegraphics[width=\linewidth]{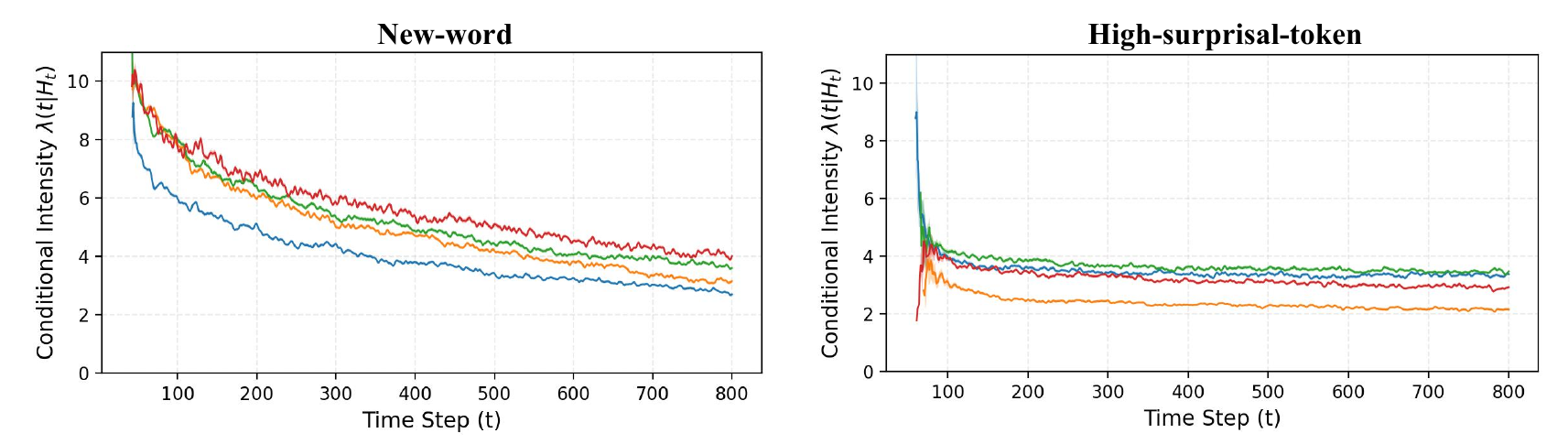}
    \caption{Writing-Prompts dataset.}
    \label{fig:5-3_writing_words_vs_entropy}
\end{subfigure}
\caption{Conditional intensity curves across event types and datasets,
    estimated from neural TPPs with semantic embeddings (LLaMA-3.1-8B). Shadow areas are 95\% bootstrap confidence intervals.}
\label{fig:5-3_words_vs_entropy_intensity}
\end{figure}

\subsection{Effects of Event Type}
\label{sec:effect_of_event_type}
We find that the conditional intensity curves of two event types exhibit different temporal patterns. 
Unlike the consistent decaying trends observed in new-word events (Figure \ref{fig:5-3_words_vs_entropy_intensity}, left column), the conditional intensity of high-surprisal-token events (Figure \ref{fig:5-3_words_vs_entropy_intensity}, right column) remains mostly stable across the sequence, except for an early oscillation at time steps $t<100$. 
The overall magnitudes of intensity are lower for the high-surprisal tokens than new words, and the visually obvious human-LLM distinction in the latter is not found in the former. 
Similar results hold for high-surprisal-token events without LLM embeddings (see Appendix \Cref{fig:entropy_without_LLM}).
High-surprisal-token events are also associated with generally higher MAE values (Table \ref{tab:TPP_mae_results}). 
We note, however, that their lower intensity implies longer inter-event intervals, which naturally inflates absolute MAE measured in time steps. Thus, to conclude whether high-surprisal-token events are inherently harder to predict requires an interval-normalized comparison, which we leave to future work. 

This asymmetry between the two event types aligns with the lossy-context surprisal theory \citep{futrell2020lossy}: human language processing operates on imperfect memory representations of the context, under which introducing a new word is disproportionately costly. Human speakers therefore preferentially reuse vocabulary established in the discourse. 
LLMs, by contrast, have nearly lossless access to the preceding context, which means retrieving a \emph{new} word is far less costly compared to humans. 
The flat curves of high-surprisal-token event intensity in human texts also empirically align with the uniform information density (UID) hypothesis \citep{jaeger2010redundancy}, which posits that language production avoids local information spikes. 
% Further supporting results can be found in \Cref{sec:appendix_all_hazard_MAE_results}.

% Intuitively, the frequency of new-word events can be considered as a rough measure of information density. However, our results reveal a more complex relationship between the two. While human texts exhibit significantly lower new-word events intensity, the difference from LLMs in high-surprisal events is not as pronounced. 

% todo: Maybe discuss on the training of LLM that causes the difference. 
% I think it may related to decoding strategies.
Finally, the reduced human-LLM difference for high-surprisal-token event intensity may itself partly arise from how the events are measured. 
New-word events are defined by a surface criterion that requires no probabilistic estimator, whereas high-surprisal-token events are defined by the output probabilities of an LLM --- an estimator that shares the same near-lossless access to context as the generator LLMs.
Therefore, the less difference in intensity curves may just reflect a blind spot of the surprisal metric. 
Using a less ``perfect'' way of computing surprisal (e.g., truncated or noised history) and then extracting events based on the imperfect surprisal could offer a direct test, which we leave to future work. 
% We also conjecture the event-type difference could also be attributed to potential biases introduced by measuring surprisal based on LM's output probabilities. For human readers, a higher frequency of new words combined with lower reuse of previous vocabulary may place significant higher cognitive load that leads to production/comprehension difficulties; however, from the perspective of an LLM estimator, a higher frequency of new words does not necessarily result in a significant change in surprisal. 
% Figure \ref{fig:5-1_intensity} may point to some systematic differences between human writing/reading processes and LLM next-token prediction.

\subsection{Classification Performances}
% \subsection{Human-LLMs Difference}
% binary: human vs. llm (gpt-4 + qwen)
% three-class or binary between models
We further investigate whether the temporal patterns of new-word occurrences distinguish human from LLM-generated texts.
We extract statistical features from the intensity curves estimated by trained neural TPPs, and feed these features into a logistic regression classifier to perform binary classification (Human vs. LLM).
50\% of the classification set is used for training and the remaining 50\% for testing.
Separate binary classifications are run for each generation model against human texts, with results reported in Table~\ref{tab:cls_main_results}.
% A full version containing other TPP architectures is provided in Appendix~\ref{sec:appendix_all_cls_results}.

% \begin{table*}[!htbp]
% \centering
% \caption{Human vs. LLM classification results based on features extracted from the intensity functions estimated by the neural TPP.}
% \label{tab:cls_main_results}
% \resizebox{0.8\textwidth}{!}{%
% \begin{tabular}{l|cc|ccc|ccc}
% \toprule
% \multirow{2}{*}{\textbf{Model}} & \multirow{2}{*}{\textbf{Accuracy}} & \multirow{2}{*}{\textbf{Macro F1}} & \multicolumn{3}{c|}{\textbf{Machine (Generated)}} & \multicolumn{3}{c}{\textbf{Human}} \\
% \cmidrule(lr){4-6} \cmidrule(lr){7-9}
% & & & Precision & Recall & F1 & Precision & Recall & F1 \\
% \midrule
% \multicolumn{9}{c}{\textbf{Speech Dataset}} \\
% \midrule
% Claude-sonnet-4.5 & 0.945 & 0.945 & 0.932 & 0.960 & 0.946 & 0.959 & 0.930 & 0.944 \\
% GPT-4.1 & 0.965 & 0.965 & 0.943 & 0.990 & 0.966 & 0.990 & 0.940 & 0.964 \\
% Qwen-plus & 0.980 & 0.980 & 0.980 & 0.980 & 0.980 & 0.980 & 0.980 & 0.980 \\
% \midrule
% \multicolumn{9}{c}{\textbf{Writing Prompt Dataset}} \\
% \midrule
% Claude-sonnet-4.5 & 0.854 & 0.854 & 0.850 & 0.858 & 0.854 & 0.857 & 0.849 & 0.853 \\
% GPT-4.1 & 0.914 & 0.914 & 0.908 & 0.921 & 0.915 & 0.920 & 0.907 & 0.913 \\
% Qwen-plus & 0.959 & 0.959 & 0.946 & 0.972 & 0.959 & 0.971 & 0.945 & 0.958 \\
% \bottomrule
% \end{tabular}
% }
% \end{table*}

\begin{table}[!htbp]
\centering
\caption{Human vs. LLM classification results based on features extracted from the intensity curves.}
\label{tab:cls_main_results}
\small
\begin{tabular}{l|ccc}
\toprule
\textbf{Model} & \textbf{Accuracy} & \textbf{Macro F1} & \textbf{AUROC} \\
\midrule
\multicolumn{4}{c}{\textbf{Speech Dataset}} \\
\midrule
Claude-sonnet-4.5 & 0.9450 & 0.9450 & 0.9913 \\
GPT-4.1 & 0.9650 & 0.9650 & 0.9932 \\
Qwen-plus & 0.9800 & 0.9800 & 0.9988 \\
\midrule
\multicolumn{4}{c}{\textbf{Writing Prompt Dataset}} \\
\midrule
Claude-sonnet-4.5 & 0.8535 & 0.8535 & 0.9263 \\
GPT-4.1 & 0.9140 & 0.9140 & 0.9719 \\
Qwen-plus & 0.9585 & 0.9585 & 0.9937 \\
\bottomrule
\end{tabular}
\end{table}

\Cref{tab:cls_main_results} shows that the features from new-word event intensity scores achieve a good classification performance. 
Qwen-Plus has the highest accuracy on both datasets, indicating that its temporal patterns diverge most from human language, which is consistent with its higher intensity curves in \Cref{fig:5-1_intensity}. Claude-Sonnet-4.5 exhibits the most ``human-like'' patterns, reflected on its lower classification performance.

% The feature importance scores under different setups are provided at \Cref{fig:feature_importance_heatmap}. Using only top-3 features with tree depth 3 (for interpretability), performance decreases negligibly (1.0\% average), demonstrating these features capture the most critical temporal characteristics. Example decision trees and scatter plots are in \Cref{fig:cls_scatter_plots_speech_llama_qwen}.
We conduct permutation feature importance analysis to identify which features contribute most. The results, shown in Appendix \ref{sec:appendix_feature_importance}, indicate that \textbf{temporal characteristics} (number of peaks and trend slope) are the most discriminative features. For local statistics, we find that the beginning segments of the intensity curves are more important, suggesting that \textbf{the initial rhythm of new-word introduction} is a key difference between human and LLM-generated texts.

Speech is consistently easier to be classified, with accuracy ranging from 0.945 to 0.980, while Writing Prompts is harder, ranging from 0.854 to 0.959.
This gap may be attributed to text \emph{genre} and LLM \emph{training}: when giving a public speech, one needs to control the rhythm of introducing new concepts for the ease of audience. This subtle design may not be well captured by LLMs; creative writing allows for more stylistic freedom, thus LLMs can better mimic human patterns. Training wise, LLMs are trained predominantly on web texts/books, lacking speech transcripts, naturally yielding larger divergence in generated speech text.

\section{Related Work}
% Meanings to CogSci
% This study proposes a novel approach for modeling the information arrival process in natural language. 
\textbf{Information dynamics in language:}
How information is distributed over time in language is a long-standing research interest in psycholinguistics. Different temporal patterns have been confirmed, including the paragraph-level steady growth of entropy \citep{genzel2003variation}, the sentence-level preference for uniform information density \citep{jaeger2010redundancy,collins2014information,meister-etal-2021-revisiting}, to the fluctuation and periodicity patterns \citep{xu2018information,tsipidi-etal-2025-harmonic,ou2025periodicity}. 
These studies primarily represent information dynamics as variation in a dense signal, and characterize it through statistical analysis. In this study, we take a discrete view of information -- rather than asking how the magnitude of an information-theoretic quantity varies over position, we model the occurrences as a context-dependent stochastic process.  
At a different level of analysis, recent work formalizes autoregressive language models as stochastic processes and studies the measure-theoretic properties such as tightness \citep{du2023measure,du2024language}, which is related to the topic of this study. 
% Other than empirical studies, there are some recent theoretical explorations that discuss the stochastic process behind language models \citep{du2023measure,du2024language}, which is related to the topic of this study. 

% Surprisal and entropy are foundational in psycholinguistics for modeling cognitive load and language processing\citep{SOULE197457, DBLP:conf/naacl/Hale01, levy2008expectation, DBLP:journals/llc/Hale16}.
\begin{comment}
Early theories like Entropy Rate Constancy (ERC) \citep{genzel2002entropy, genzel2003variation, verma-etal-2023-revisiting} and Uniform Information Density (UID) \citep{jaeger2006speakers,jaeger2010redundancy,collins2014information,meister-etal-2021-revisiting} suggest humans distribute information evenly.
However, recent studies reveal that natural discourse is not strictly uniform; instead, it exhibits harmonic structures and periodic entropy fluctuations  \citep{tsipidi-etal-2024-surprise, tsipidi-etal-2025-harmonic, ou2025periodicity}.
Spectral and Fourier analysis of cross-entropy sequences \citep{yang2023face, DBLP:conf/emnlp/LiuLXWYY25} captures frequency-domain differences, highlighting that temporal information dynamics are crucial for distinguishing human language from LLM outputs.
\end{comment}
% (UID and the deterministic/stochastic discussion are de-emphasized here; the full treatment lives in Related Work.)

\paragraph{TPPs and language models:}
Early work modeled the arrival of new words in text using Poisson processes with slowly evolving rates~\citep{badalamenti1994poisson,badalamenti1994lawful,strand1997systems,badalamenti2001speech}.
Modern neural TPPs improve the flexibility of event-sequence modeling by learning representations of event history and parameterizing the intensity with neural networks~\citep{du2016recurrent,mei2017_6463c884,omi2019fully,shchur2020_00ac8ed3}.
More recently, TPP-LLM frameworks have used LLMs to encode event descriptions or process tokenized event sequences, improving the prediction of real-world events~\citep{liu2024tpp,kong2025language}.
In these studies, however, language primarily serves as a representation of events; language process per se is not under investigation.
\section{Conclusions}
We propose to model the arrival time of information-bearing events in natural language using neural temporal point processes (TPPs). We define ``events'' as new words and high-surprisal tokens. The main findings are: neural TPPs can predict the arrival time of next event with decently good accuracies. % Taking the inter-event intervals and the contextual representations as input, 
The learned event intensity curves of new words reveal a significant difference between human-produced and LLM-generated texts. 
Our results indicate the unique human cognitive mechanism that makes the new-word production more costly than LLMs. 
The less pronounced difference on suprisal-based events also points out the bias of using surprisal as an information metric. 
The proposed approach will be potentially useful for developing more flexible natural language generation and test-time scaling techniques, by extending LLMs' token-level predicting capabilities to timing prediction of important tokens.
% while the LLM curves have overall higher intensity and decay slower. The event intensity curves can serve as useful features to accurately distinguish human vs. LLM generations with simple classifiers.

% \clearpage
\section{Limitations}
In this work, we use TPP models to model event sequences on a continuous time axis. However, word-level and token-level events are inherently discrete, occurring only at integer points. Therefore, continuous TPP models may complicate the problem for the events we study in language. 
However, after trying several discrete methods for modeling TPP intensity functions, we found that they did not yield better performance and often exhibited instability during training. We believe that discrete TPP models may be a sound alternative, but further design and experimentation are needed to make them truly applicable. 

% Limits of event types, only two types are examined now.
Another limitation concerns the scope of our event definitions. We examine only new-word and high-surprisal-token events, which capture lexical novelty and probabilistic unexpectedness, respectively. However, many other linguistic phenomena may also be viewed as information-bearing events, including boundaries in textual structure marked by commas or sentence-final punctuation, as well as occurrences of particular lexical categories such as nouns and verbs. Extending neural TPPs to a broader range of linguistic events can provide more room for further investigation into the organization and properties of text.

% \section*{Author Contributions}

% \section*{Acknowledgments}

% \section*{Ethics Statement}
% Authors can add an optional ethics statement to the paper.
% For papers that touch on ethical issues, this section will be evaluated as part of the review process. The ethics statement should come at the end of the paper. It does not count toward the page limit, but should not be more than 1 page.

\bibliography{custom}

\clearpage
\appendix

\section{Mathematical Basics of TPPs}
\label{sec:A_tpps_math}

\subsection{Proof of the conditional probability density function}
\label{sec:A1-1_proof_PDF}
\textbf{Claim of \Cref{eq:next_event_PDF}.} Given the event history $\mathcal{H}_{t_{i}}=\{t_1,t_2,\dots,t_i\}$, the probability density function of the next event occurring at time $t_{i+1}$ is:
\begin{equation*}
\begin{split}
    p(t_{i+1} \mid \mathcal{H}_{t_{i+1}}) &= \\
    \lambda(t_{i+1} \mid &\mathcal{H}_{t_{i+1}}) \exp\left(-\int_{t_i}^{t_{i+1}} \lambda(t \mid \mathcal{H}_t)\, dt\right).
\end{split}
\end{equation*}

\paragraph{Step 1.}
Define survival function $S(t)$ as the probability of ``no event occurs in $[t_i,t)$'', $S(t) = P\big(\text{no event in } [t_i, t) \,\big|\, \mathcal{H}_{t_i}\big)$. $S(t)$ is monotonically non-increasing, and by construction, $S(t_i)=1$.

By definition of the conditional intensity function (\Cref{eq:intensity_func}), the probability of ``one event in $[t,t+\Delta)$'' conditioned on ``no event in $[t_i,t)$'' is: 

\begin{equation*}
\begin{split}
    P\big(\text{one event in } [t, t+\Delta) &\,\big|\,\mathcal{H}_t\big) \\
    = \lambda(t \mid \mathcal{H}_t)\,\Delta& + o(\Delta).
\end{split}
\end{equation*}
, in which $o(\Delta)$ is a higher order term s.t. $\lim_{\Delta\to 0} \frac{o(\Delta)}{\Delta}=0$. The key observation is that the condition ``no event in $[t_i,t)$'' together with $\mathcal{H}_{t_i}$ leads to exactly the history $\mathcal{H}_t$. Therefore, 

\begin{equation*}
    \begin{split}
        P\big(\text{no event in } [t, t+\Delta) &\,\big|\,\mathcal{H}_t\big) \\
    = 1 - \lambda(t \mid \mathcal{H}_t)\,\Delta& + o(\Delta).
    \end{split}
\end{equation*}

\paragraph{Step 2.}
Applying the multiplication rule of probability:

\begin{equation*}
\begin{split}
    S(t+\Delta) = & P\big(\text{no event in } [t_i, t) \,\big|\, \mathcal{H}_{t_i} \big) \cdot \\
    & P\big(\text{no event in } [t, t+\Delta) \,\big|\, \mathcal{H}_t\big) \\
    = & S(t)\,\big[1 - \lambda(t \mid \mathcal{H}_t)\,\Delta + o(\Delta)\big].
\end{split}
\end{equation*}

Rearranging into the form of an ODE:

\begin{equation*}
\begin{split}
    & \frac{S(t+\Delta) - S(t)}{\Delta} = - \lambda(t \mid \mathcal{H}_t)\,S(t) + \frac{o(\Delta)}{\Delta} \\
    & \Rightarrow \lim_{\Delta \to 0} \frac{S(t+\Delta) - S(t)}{\Delta} = \frac{d\,S(t)}{dt} \\
    & = - \lambda(t \mid \mathcal{H}_t)\,S(t).
\end{split}
\end{equation*}

Then solve this first-order linear ODE by integrating from $t_i$ to $t$:

\begin{equation*}
\begin{split}
    & \frac{d}{dt}\log\,S(t) = - \lambda(t \mid \mathcal{H}_t)\\
    & \Rightarrow \log\,S(t) - \log\,S(t_i) = -\int_{t_i}^{t} \lambda(s \mid \mathcal{H}_s)\,ds.
\end{split}
\end{equation*}

Since $S(t_i)=1$, we have: 

\begin{equation*}
\begin{split}
    & \log S(t) = -\int_{t_i}^{t} \lambda(s \mid \mathcal{H}_s)\,ds \\
    & \Rightarrow S(t) = \exp\left(-\int_{t_i}^{t} \lambda(s \mid \mathcal{H}_s)\,ds\right).
\end{split}
\end{equation*}

\paragraph{Step 3.}
By definition, $S(t)$ is the probability of ``no event in $[t_i,t)$''. Then the probability of ``next event $t_{i+1}$ occurs in $[t_i,t)$'' is $1-S(t)=F(t)$, which is the cumulative  distribution function that describes $p(t_{i+1}<t|\mathcal{H}_{t_i})$. Hence its density is:

\begin{equation*}
\begin{split}
    p(t|\mathcal{H}_{t_i}) & = \frac{dF(t)}{dt}=-\frac{dS(t)}{dt}=\lambda(t \mid \mathcal{H}_t)\,S(t)\\
    & = \lambda(t \mid \mathcal{H}_t)\exp\left(-\int_{t_i}^{t} \lambda(s \mid \mathcal{H}_s)\,ds\right).
\end{split}
\end{equation*}

Evaluating at $t=t_{i+1}$ yields \Cref{eq:next_event_PDF}. $\square$

\subsection{Formulating the loss function for learning neural TPPs}
\label{sec:A1-2_NTPP_loss}
\paragraph{Claim.} The loss function for learning a neural TPP with a cumulative hazard network $\Phi$ can be expressed as the negative log-likelihood of the event time data $\{t_i\}^N_{i=1}$ scattered over $[0,T]$: 
\begin{equation}
\begin{split}
    \mathcal{L}_{\text{NTPP}} &= -\log L(\{t_i\}^N_{i=1}) = \\
    & \sum_{i} \Bigl[ \log \Bigl\{ \frac{\partial}{\partial\tau}\Phi(\tau=t_{i+1}-t_i\mid \bm{h}_i) \Bigr\} \\
    & - \Phi(t_{i+1} - t_i\mid \bm{h}_i )  \Bigr].
\end{split}
\label{eq:AppA-2_NTPP_loss}
\end{equation}

\paragraph{Proof.}
Applying the probability density function \Cref{eq:next_event_PDF} to the event sequence $\{t_i\}^N_{i=1}$, we have the probability of observing this sequence:
\begin{equation*}
    p(\{t_i\}) = \prod_{i}\lambda(t_i|\mathcal{H}_{t_i})\exp\biggl\{ -\int_0^T \lambda(t \mid \mathcal{H}_t)\, dt\biggr\}.
% \label{eq:AppA-2_seq_prob}
\end{equation*}

Take the logarithm of both sides and substitute in the hazard $\lambda(t|\mathcal{H}_t) = \phi(t-t_i\mid \bm{h}_i)$:

\begin{equation}
\begin{split}
    & \log L(\{t_i\}) = \\
    & \sum_{i}\biggl[ \log\phi(t_{i+1} - t_i\mid \bm{h}_i) - \int_{0}^{t_{i+1}-t_i} \phi(s \mid\bm{h}_i)\,ds\biggr].
\end{split}
\label{eq:AppA-2_logliki_hazard}
\end{equation}

Use the definition of the cumulative hazard function \Cref{eq:cumulative_hazard_def}: $\Phi(\tau|\bm{h}_i) = \int_0^\tau \phi(s|\bm{h}_i) \, ds$. Substituting it into \Cref{eq:AppA-2_logliki_hazard} yields \Cref{eq:AppA-2_NTPP_loss}. $\square$

\section{Datasets and Language Models}
\label{sec:appendix_dataset}

\subsection{Dataset Details}
% dataset         subset                  texts    total_words  avg_words      min      max
% ------------------------------------------------------------------------------------------
% speech          Human                     200         309958    1549.79      661     2875
% speech          Claude-sonnet-4.5         200         324422    1622.11      961     3058
% speech          GPT-4.1                   200         347851    1739.26     1051     2902
% speech          Qwen-plus                 200         342591    1712.95      994     2776
% speech          TOTAL                     800        1324822    1656.03      661     3058
% ------------------------------------------------------------------------------------------
% writing_2000    Human                    2000        2337315    1168.66      807     1501
% writing_2000    Claude-sonnet-4.5        2000        2337204    1168.60      911     1816
% writing_2000    GPT-4.1                  2000        2567079    1283.54      762     1868
% writing_2000    Qwen-plus                2000        3033722    1516.86      939     2342
% writing_2000    TOTAL                    8000       10275320    1284.41      762     2342
% ------------------------------------------------------------------------------------------
% writing_10000   Human                   10000       11670773    1167.08      750     1512
% writing_10000   TOTAL                   10000       11670773    1167.08      750     1512
% ------------------------------------------------------------------------------------------

\paragraph{Speech}
For the transcripts of human speeches, we select 200 speeches from the \href{https://www.americanrhetoric.com/speechbank.htm}{American Rhetoric} speech bank. For each speech, the length is controlled to be between 800 and 2500 words, ensuring the texts are sufficiently long for robust temporal pattern analysis.

\paragraph{Writing Prompt}
For the human-written stories, we use the \texttt{Writing-Prompts} dataset \citep{writingprompts}, which is a collection of stories written by human writers in response to various prompts.
We filter the dataset to select 12000 stories, each of which has a length between 1000 and 2000 words, a prompt that is long enough (at least 20 words), and does not contain long segments of non-English characters. From this filtered set, we randomly select 2000 stories for temporal pattern analysis, and retain the remaining 10000 stories for the classification task.

\subsection{Machine Text Generation}
\label{sec:appendix_prompting}

We experiment with three state-of-the-art LLMs: Claude-Sonnet-4.5, GPT-4.1 and Qwen-Plus. To establish a balanced comparative dataset, we prompt these LLMs to generate texts for the 200 Speeches and 2000 stories used in our temporal pattern analysis. For each generated text, we instruct the LLMs to closely match the length of the corresponding human-authored text. %The detailed prompting templates are provided in \Cref{sec:appendix_prompting}.

% \subsubsection{Speech Generation Template}
\paragraph{Speech Generation Template}
\begin{center}
\fbox{\begin{minipage}{.9\linewidth}
\vspace{2mm}
Please write a long speech of about \{length\} words on the topic of \{topic\}, using solemn and rhetorically refined language (avoid tired or clichéd expressions), and include numerous concrete examples throughout.
\vspace{2mm}
\end{minipage}}
\end{center}

% \subsubsection{Writing Prompt Story Generation Template}
\paragraph{Writing Prompt Story Generation Template}
\begin{center}
\fbox{\begin{minipage}{.9\linewidth}
\vspace{2mm}
Write a creative and engaging story based on the following premise. Ensure the story directly explores the given setup and develops it into a complete narrative.

Premise: \{prompt\}

Additional Guidelines:

1. Let the premise guide the story's direction, but feel free to develop characters, conflict, and resolution organically.

2. The story should be self-contained and have a satisfying narrative flow.

3. Aim for a story length of approximately \{length\} words. Please be mindful of this target length as you write.

Story:
\vspace{2mm}
\end{minipage}}
\end{center}

\subsection{Embedding Models}
% To capture the contextual semantic information ($e_i$) required by our neural TPP architectures, we utilize two distinct, lightweight pre-trained language models: LLaMA-3.2-1B and Qwen-3-0.6B. We select these small-parameter models from two different families to maintain computational efficiency and to mitigate potential architectural biases.
% we used three different models: LLaMA-3.1-8B, LLaMA-3.2-1B, and Qwen-3-0.6B. These choices include different families, and different scales from the same family.
To extract the contextual semantic embeddings ($e_i$), we primarily utilize LLaMA-3.1-8B \citep{grattafiori2024llama3herdmodels} for the core experiments reported in the main text. 
To rigorously evaluate the robustness of our Neural TPP framework across different model architectures and parameter scales, we conduct supplementary experiments using LLaMA-3.2-1B\citep{grattafiori2024llama3herdmodels} and Qwen-3-0.6B\citep{yang2025qwen3}.
Detailed results and temporal dynamics for these lightweight models are provided in \Cref{sec:appendix_Lightweight_model}.

\section{Implementation Details}
\label{sec:appendix_Implementation}

\subsection{Dequantization for Discrete-time}
\label{sec:dequantization}
Because our linguistic events are defined over a discrete temporal domain (i.e., integer-based wait times between events), whereas neural TPPs are originally designed for continuous-time processes, we apply dequantization during training \citep{DBLP:journals/corr/TheisOB15, DBLP:conf/icml/HoCSDA19}.
Specifically, we inject a small amount of zero-mean Gaussian noise with a small standard deviation ($\sigma=0.05$), truncated to $[-0.5, 0.5]$, into the discrete target inter-event intervals.
This ensures that approximately 95\% of the noise falls within $[-0.1, 0.1]$, successfully breaking the mathematical singularities (infinite density spikes) at exact integer points without blurring the boundaries between adjacent positions.
This adaptation prevents numerical instability and allows the continuous intensity function to smoothly fit the discrete text generation process.

\subsection{Semantic Embeddings and PCA}
\label{sec:PCA}
 The raw semantic embeddings extracted from the pre-trained LLMs initially have a high dimensionality (4096 for LLaMA-3.1-8B, 2048 for LLaMA-3.2-1B and 1024 for Qwen-3-0.6B).
To enhance computational efficiency while preserving contextual semantic information, the semantic embeddings extracted from the pre-trained LLMs are projected onto a lower-dimensional space using Principal Component Analysis (PCA) \citep{abdi2010principal}.
\begin{itemize}[leftmargin=10pt]
    \item For the Speech dataset, the reduced PCA dimension is set to 64 across all architectures.
    \item For the standard analysis on the Writing Prompt dataset, the PCA dimension is set to 128.
    \item For the larger classification task on the Writing Prompt dataset, the PCA dimension is expanded to 256 to accommodate the larger data scale.
\end{itemize}

\subsection{Model Architectures and Training Configurations}
\label{sec:Configurations}
All neural TPP models are optimized to minimize the negative log-likelihood (NLL) of event occurrences using the Adam optimizer \citep{DBLP:journals/corr/KingmaB14}.
To ensure reproducibility and stable training, we apply an initial learning rate of $1 \times 10^{-3}$, and a dropout rate of 0.1 across all setups.
The Cumulative Hazard Function Network consistently utilizes 2 layers with a hidden size of 64 across all standard setups (scaled to 128 for the larger classification task). %We dynamically scale the model capacity based on the dataset complexity and task volume:
\paragraph{RNN-based neural TPP}
We set the sliding window size of recent historical inter-event intervals to $w=40$, aligning with the maximum length explored by \citet{omi2019fully} to ensure that sufficient historical context is captured.
\begin{itemize}[leftmargin=10pt]
\item For the Speech dataset, we set the RNN hidden size and positional encoding dimension to 32, and train with a batch size of 128.
\item For the Writing Prompt dataset, the RNN hidden size and positional encoding dimension are increased to 64, with a batch size of 512.
\item For the classification task, the model capacity is further expanded: the RNN hidden size and positional encoding dimension are set to 128, with a batch size of 2048.
\end{itemize}

\paragraph{Transformer-based neural TPP}
For the architectural variation explored in \Cref{sec:tansformer-based}, the Transformer-based models process the entire sequence history using stacked causal attention layers.
\begin{itemize}[leftmargin=10pt]
\item For the Speech dataset, the model employs a dimension ($d_{model}$) of 16, and a feed-forward inner dimension of 32. It uses 2 attention heads and 2 layers, trained with a batch size of 4.
\item For the Writing Prompt dataset, the capacity is scaled up: $d_{model} = 32$,  and inner dimension is 64, maintaining 2 heads and 2 layers, with a batch size of 8.
% \item For the classification task, the capacity is significantly expanded: $d_{model} = 64$, and inner dimension is 128, utilizing 4 heads and 4 layers, with a batch size of 64.
\end{itemize}

\section{Further Supporting Results}

\label{sec:further_supporting_results}

\subsection{Conventional TPPs Fitting}
\label{sec:Conventional_TPPs_fitting}
Table \ref{tab:conventional_tpp_results} presents the detailed goodness-of-fit metrics for the Conventional TPP model when linguistic events are defined as new words. The parametric model demonstrates a strong fit across all text sources (human and LLM-generated) in both the Speech and Writing Prompt datasets. Specifically, the Residual Mean Square (RMS) remains consistently low (averaging approximately 0.0015), and the correlation (R) between the fitted and observed data is exceptionally high (consistently exceeding 0.96, with max values often reaching above 0.99). This fitting result closely aligns with the research by \citet{badalamenti1994poisson}.
% These robust fitting metrics validate the use of the Conventional TPP as a reliable baseline.

\begin{table*}[!htbp]
\centering
\caption{Goodness-of-fit evaluation of the Conventional TPP model when events are defined as new words. The metrics include the Residual Mean Square (RMS) of the regression and the correlation ($R$) between fitted and observed data.}
\label{tab:conventional_tpp_results}
\resizebox{0.85\textwidth}{!}{%
\begin{tabular}{l|ccc|ccc}
\toprule
\multirow{2}{*}{\textbf{Model}} & \multicolumn{3}{c|}{\textbf{RMS} $\downarrow$} & \multicolumn{3}{c}{\textbf{R} $\uparrow$} \\
\cmidrule(lr){2-4} \cmidrule(lr){5-7}
& Mean & Min & Max & Mean & Min & Max \\
\midrule
\multicolumn{7}{c}{\textbf{Speech Dataset}} \\
\midrule
Human             & 0.001191 & 0.000191 & 0.004748 & 0.991702 & 0.963751 & 0.997403 \\
Claude-Sonnet-4.5 & 0.001753 & 0.000364 & 0.005538 & 0.990342 & 0.974717 & 0.996734 \\
GPT-4.1           & 0.001415 & 0.000199 & 0.004075 & 0.990706 & 0.974341 & 0.997319 \\
Qwen-Plus         & 0.001392 & 0.000191 & 0.005521 & 0.989525 & 0.974030 & 0.997438 \\
\midrule
\multicolumn{7}{c}{\textbf{Writing Prompt Dataset}} \\
\midrule
Human             & 0.001342 & 0.000148 & 0.008940 & 0.992903 & 0.976296 & 0.998898 \\
Claude-Sonnet-4.5 & 0.001563 & 0.000142 & 0.011887 & 0.993475 & 0.975005 & 0.998647 \\
GPT-4.1           & 0.001538 & 0.000214 & 0.027781 & 0.993070 & 0.960826 & 0.998924 \\
Qwen-Plus         & 0.001538 & 0.000139 & 0.012236 & 0.992548 & 0.963879 & 0.999228 \\
\bottomrule
\end{tabular}%
}
\end{table*}

\subsection{Supplementary Conditional Intensity Curves for Main Experiments}
\label{sec:D2-intensity_curves_ntpp}
\begin{figure*}[!htbp]
    \centering
    \includegraphics[width=0.85\linewidth]{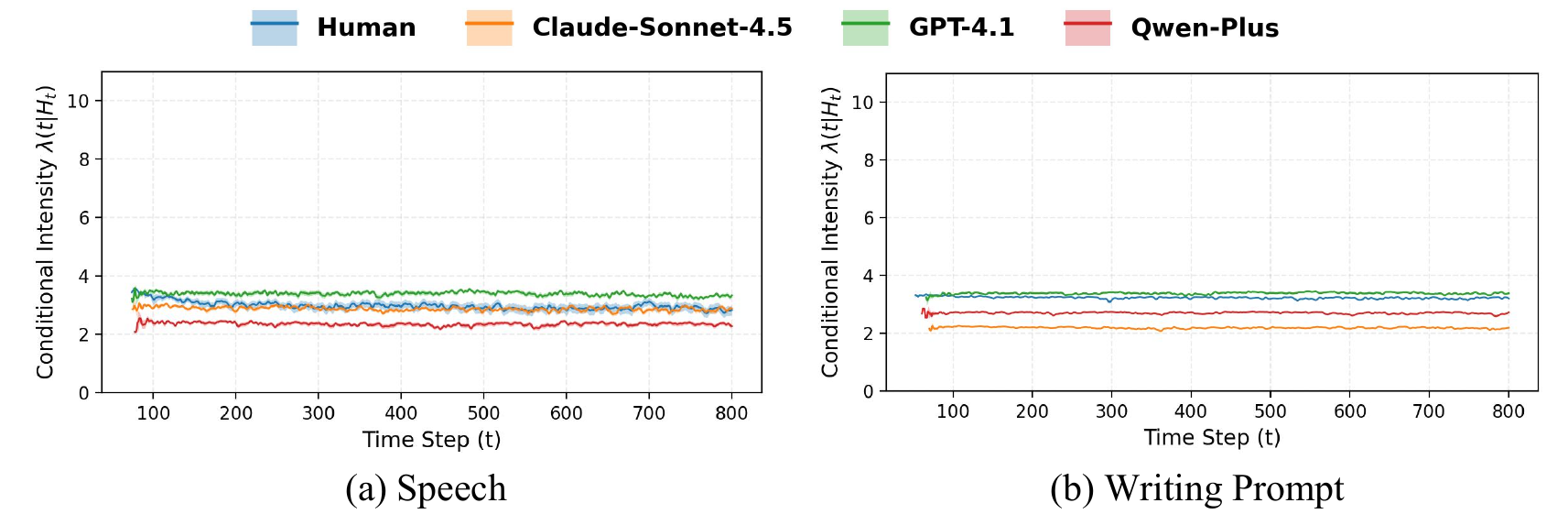}
    \caption{Estimated conditional intensity functions with LLaMA-3.1-8b-derived high-surprisal-token events on the both datasets. Both subplots utilize RNN-based neural TPPs without LLM embeddings (w/o LLM). Shadow areas are 95\% bootstrap confidence intervals.}
    \label{fig:entropy_without_LLM}
\end{figure*}

As discussed in \Cref{sec:effect_of_event_type} of the main text, this section provides supplementary visualization for the temporal dynamics of linguistic events. \Cref{fig:entropy_without_LLM} displays the conditional intensity curves for high-surprisal-token events across both datasets when the neural TPPs rely exclusively on temporal features without LLM semantic embeddings. As shown in the figure, the temporal dynamics—specifically the flat, stable trajectory for high-surprisal-token events—are highly consistent with the primary trends analyzed in the main text.

\begin{figure*}[!htbp]
    \centering
    \includegraphics[width=0.85\linewidth]{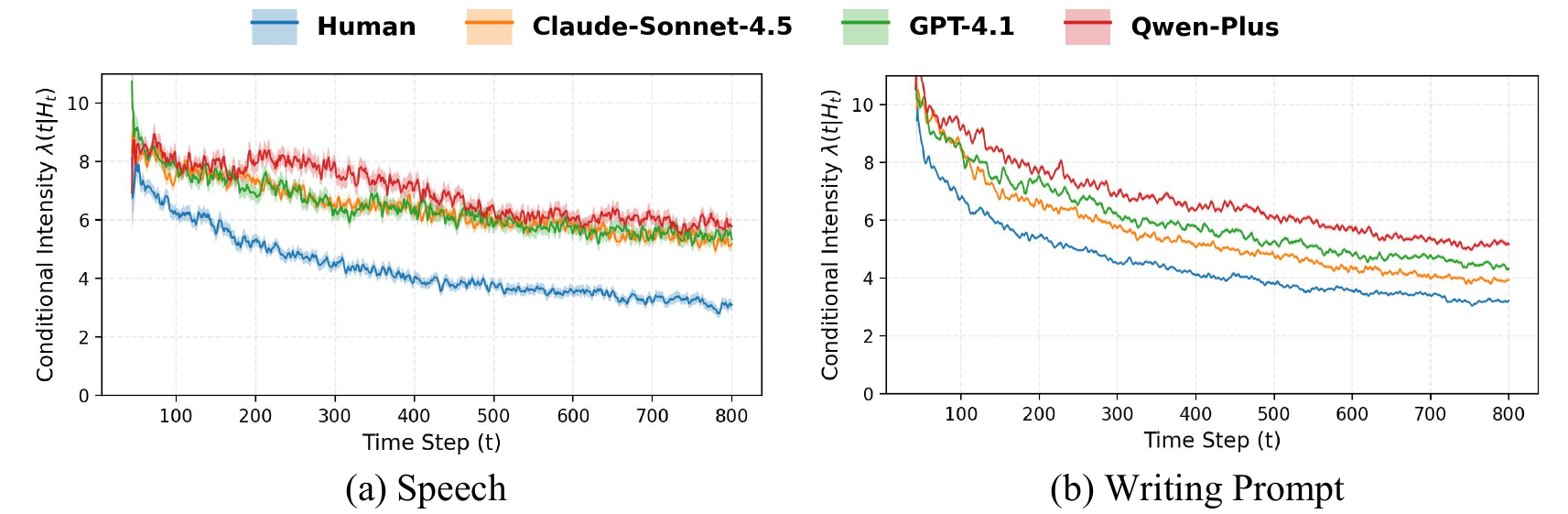}
    \caption{Estimated conditional intensity functions with new-word events \textbf{without lemmatization} on the both datasets. Both subplots utilize RNN-based neural TPP incorporating semantic embeddings extracted via LLaMA-3.1-8B. Shadow areas are 95\% bootstrap confidence intervals.}
    \label{fig:D-3_word_without_lemmatization}
\end{figure*}

\subsection{The Impact of Lemmatization on New-Word Events}
\label{sec:D4-new_words_lemmatization}

In \Cref{sec:def_events}, we defined new-word events by lemmatizing nouns and verbs. This selective preprocessing is intended to avoid counting common inflectional variants as entirely new lexical items:
\begin{itemize}[leftmargin=10pt]
\item \textbf{Nouns and Verbs (Lemmatized):} Morphological variants of nouns and verbs (e.g., pluralization like "cat" vs. "cats", or tense changes like "run" vs. "ran") are predominantly inflectional\citep{anderson1982where}. 
They are dictated by strict grammatical constraints and refer to the exact same core conceptual entities or actions. 
Following \citet{badalamenti1994poisson, badalamenti1994lawful}, who lemmatized verbs, we also lemmatize nouns so that these inflectional variants are counted consistently.

% \item \textbf{Adjectives and Adverbs (Not Lemmatized):} Variations in adjectives and adverbs (such as comparatives and superlatives, e.g., "good" vs. "better"), while also involving inflectional morphology, serve a fundamentally different purpose. The introduction of comparative forms in discourse is rarely driven purely by syntactic requirements. According to degree semantics, gradable modifiers map entities to degrees on a semantic scale\citep{kennedy1999projecting}, making them a deliberate injection of new semantic information rather than just a grammatical shift.
\item \textbf{Adjectives and Adverbs (Not Lemmatized):} We retain the surface forms of adjectives and adverbs, including comparative and superlative forms such as ``good'' and ``better''.
While these forms are morphologically related, they convey distinct semantic information. Degree semantics distinguishes them because they express different relations on a semantic scale\citep{kennedy1999projecting}. We therefore treat them as distinct forms in our event extraction procedure.
\end{itemize}
To assess whether the observed human--LLM differences are sensitive to this preprocessing choice, we repeat the analysis using new-word events extracted without lemmatizing any words.

As illustrated in \Cref{fig:D-3_word_without_lemmatization}, the trends of the resulting conditional intensity curves remain consistent with our main results (\Cref{fig:5-3_words_vs_entropy_intensity}, left column).
In particular, the human intensity curves generally remain below those of Claude-Sonnet-4.5, GPT-4.1, and Qwen-Plus. This comparison suggests that the observed difference is robust and not merely an artifact of the lemmatization strategy.

\subsection{Results on Lightweight Language Models}
\label{sec:appendix_Lightweight_model}

To ensure our findings regarding the temporal dynamics of linguistic events are not biased by the choice of semantic extractor, we replicated the intensity estimation using lightweight models, specifically LLaMA-3.2-1B and Qwen-3-0.6B. For high-surprisal-token events, the corresponding token-level surprisal values were also computed using these lightweight models respectively.

\Cref{tab:different_model_mae_results} provides the Mean Absolute Error (MAE) for the neural TPP models across different event types and text sources. 
For new-word events, the MAE differences across various LLMs are marginal, suggesting that the predictability of these events is robust to the parameter scale of the semantic extractor.
In contrast, high-surprisal-token events exhibit noticeably larger MAE variations across LLMs.
% This discrepancy may stem from the dual role of LLMs in this specific setup: they act as both contextual semantic extractors and entropy evaluators.
This discrepancy may arise because the different LLMs, when acting as high-surprisal-token evaluators, inherently possess distinct probability distributions. Consequently, the extracted sequences of high-surprisal-token events naturally diverge, introducing larger fluctuations in the predictive performance of the neural TPP models.
This empirically supports the discussion in \Cref{sec:effect_of_event_type} regarding the potential biases introduced by using an LLM's output probabilities to measure surprisal.
% The results confirm that the observed temporal patterns are robust across different embedding models, maintaining highly consistent predictive capabilities.

\begin{figure*}[!htbp]
\centering
\begin{subfigure}[b]{0.85\textwidth}
    \centering
    \includegraphics[width=\linewidth]{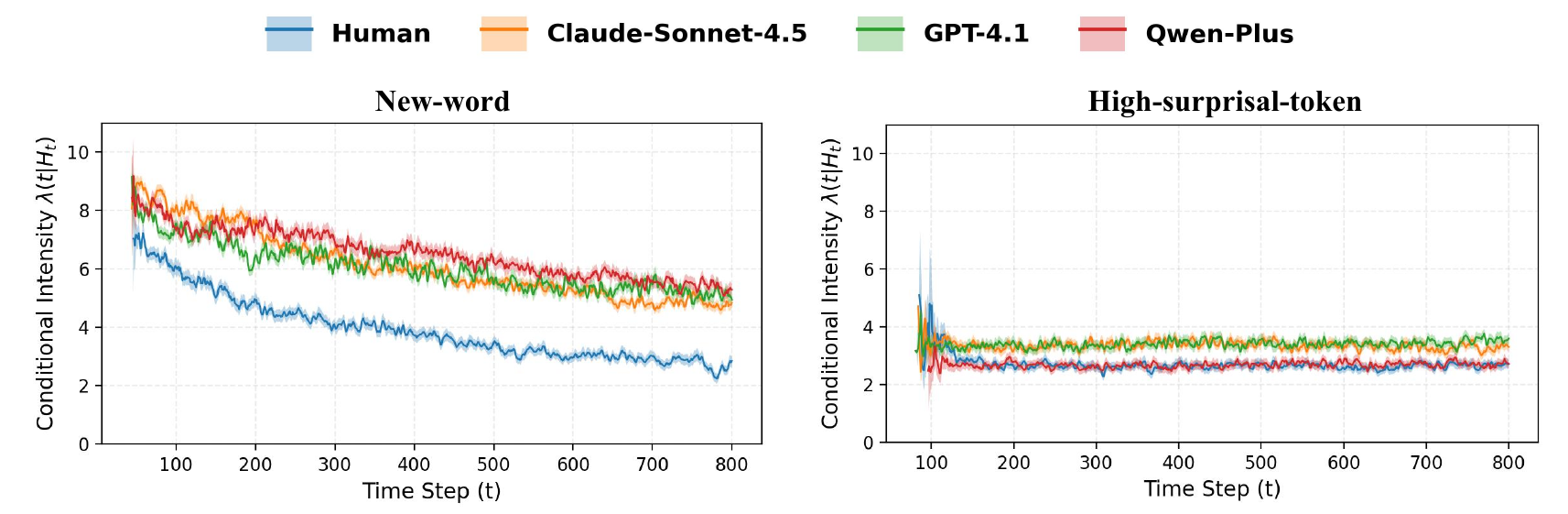}
    \caption{Speech dataset.}
    \label{fig:D-4_llama_1b_intensity_speech}
\end{subfigure}
\hfill
\begin{subfigure}{0.85\textwidth}
    \centering
    \includegraphics[width=\linewidth]{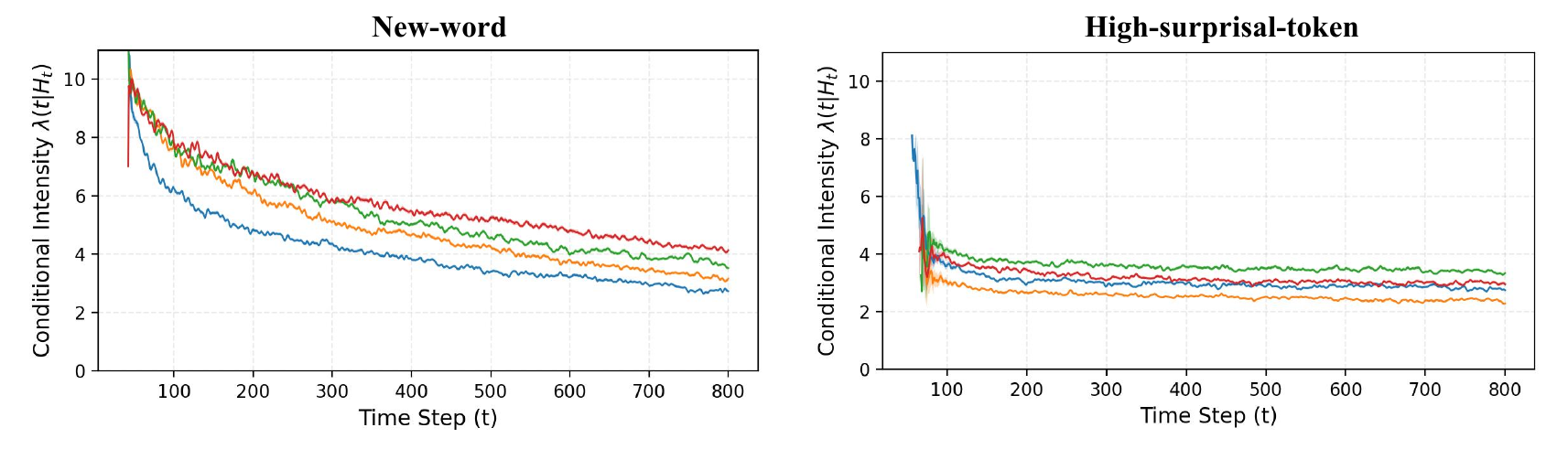}
    \caption{Writing-Prompts dataset.}
    \label{fig:D-4_llama_1b_intensity_writing}
\end{subfigure}
\caption{Estimated conditional intensity functions across different linguistic event types.
    All subplots utilize RNN-based neural TPP incorporating semantic embeddings extracted via \textbf{LLaMA-3.2-1B}. Shadow areas are 95\% bootstrap confidence intervals.}
\label{fig:D-4_llama-1b_intensity}
\end{figure*}

\begin{figure*}[!htbp]
\centering
\begin{subfigure}[b]{0.8\textwidth}
    \centering
    \includegraphics[width=\linewidth]{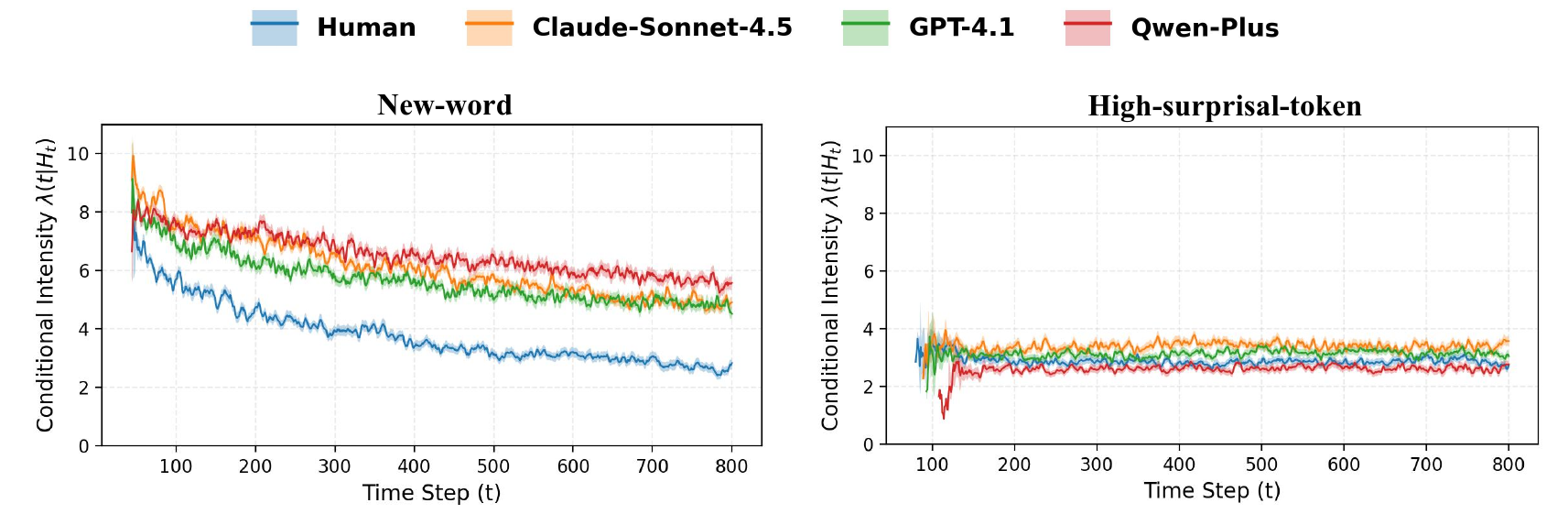}
    \caption{Speech dataset.}
    \label{fig:D-4_qwen-0.6b_intensity_speech}
\end{subfigure}
\hfill
\begin{subfigure}{0.8\textwidth}
    \centering
    \includegraphics[width=\linewidth]{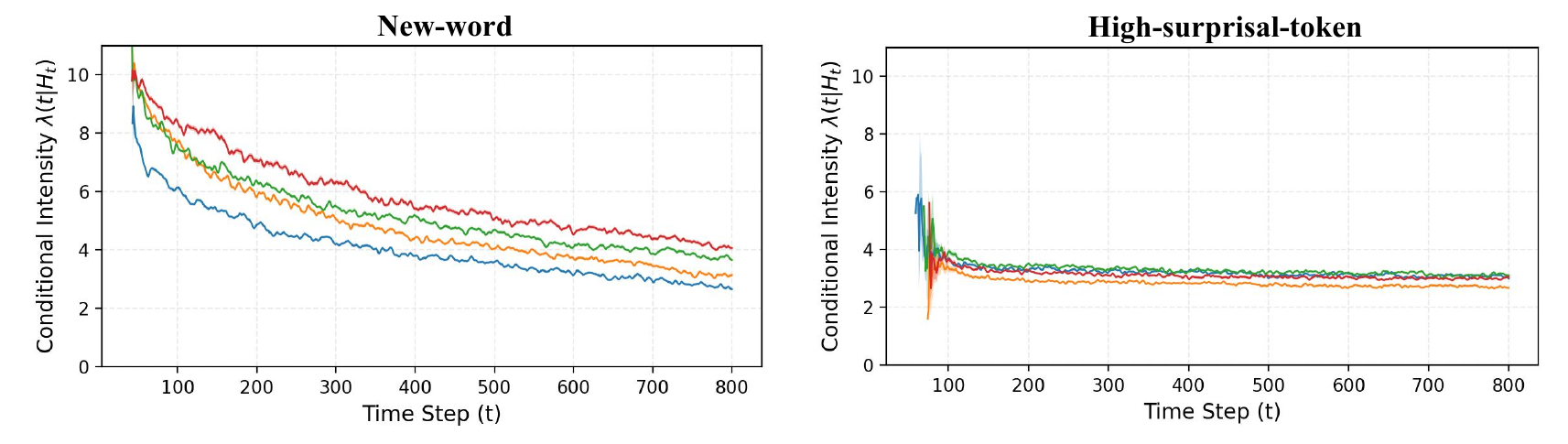}
    \caption{Writing-Prompts dataset.}
    \label{fig:D-4_qwen-0.6b_intensity_writing}
\end{subfigure}
\caption{Estimated conditional intensity functions across different linguistic event types.
    All subplots utilize RNN-based neural TPP incorporating semantic embeddings extracted via \textbf{Qwen-3-0.6B}. Shadow areas are 95\% bootstrap confidence intervals.}
\label{fig:D-4_qwen-0.6b_intensity}
\end{figure*}

Nevertheless, the overall temporal dynamics captured by the models remain highly consistent. 
\Cref{fig:D-4_llama-1b_intensity} and \Cref{fig:D-4_qwen-0.6b_intensity} show the conditional intensity functions utilizing these lightweight embeddings, which align closely with the LLaMA-3.1-8B results.
% The overall trajectory trends in these figures are highly consistent with those observed using LLaMA-3.1-8B. 
Specifically, the intensity of new-word events exhibits a continuous decay over time, with LLMs persistently maintaining a higher generation rate of new words than humans. 
In straightforward contrast, the intensity curves for high-surprisal tokens remain remarkably flat and stable across the entire generation sequence.

% \begin{table*}[!htbp]
% \centering
% \caption{Comparison of predicting performance (MAE$\downarrow$) for RNN-based neural TPPs with semantic embeddings from LLaMA-3.1-8B, LLaMA-3.2-1B, and Qwen-3-0.6B.}
% \label{tab:different_model_mae_results}
% \resizebox{0.8\textwidth}{!}{ 
% \begin{tabular}{ll|l|ccc}
% \toprule
% \textbf{Dataset} & \textbf{Event Type} & \textbf{Text Source} & \textbf{\makecell{w/ LLaMA-\\3.1-8B}} & \textbf{\makecell{w/ LLaMA-\\3.2-1B}} & \textbf{\makecell{w/ Qwen-\\3-0.6B}} \\
% \midrule

% \multirow{8}{*}{\textbf{Speech}} & \multirow{4}{*}{\textbf{New-word}} 
% & Human & 1.726 & 1.602 & 1.611 \\
% & & Claude-Sonnet-4.5 & 1.019 & 0.978 & 0.976 \\
% & & GPT-4.1 & 1.009 & 0.964 & 0.971 \\
% & & Qwen-Plus & 0.859 & 0.830 & 0.837 \\
% \cmidrule(lr){2-6}

% & \multirow{4}{*}{\textbf{\makecell{High-surprisal-\\token}}} 
% & Human & 1.780 & 2.011 & 1.804 \\
% & & Claude-Sonnet-4.5 & 1.624 & 1.540 & 1.532 \\
% & & GPT-4.1 & 1.322 & 1.456 & 1.581 \\
% & & Qwen-Plus & 2.003 & 1.817 & 1.989 \\
% \midrule

% \multirow{8}{*}{\textbf{\makecell{Writing \\ Prompt}}} & \multirow{4}{*}{\textbf{New-word}} 
% & Human & 1.462 & 1.347 & 1.361 \\
% & & Claude-Sonnet-4.5 & 1.265 & 1.171 & 1.179 \\
% & & GPT-4.1 & 1.118 & 1.045 & 1.054 \\
% & & Qwen-Plus & 1.075 & 1.015 & 1.021 \\
% \cmidrule(lr){2-6}

% & \multirow{4}{*}{\textbf{\makecell{High-surprisal-\\token}}} 
% & Human & 1.569 & 1.736 & 1.656 \\
% & & Claude-Sonnet-4.5 & 2.285 & 1.964 & 1.860 \\
% & & GPT-4.1 & 1.413 & 1.452 & 1.554 \\
% & & Qwen-Plus & 1.799 & 1.569 & 1.771 \\

% \bottomrule
% \end{tabular}
% }
% \end{table*}

\begin{table*}[!htbp]
\centering
\caption{Comparison of predicting performance (MAE$\downarrow$) for RNN-based and Transformer-based neural TPPs with semantic embeddings from LLaMA-3.1-8B, LLaMA-3.2-1B, and Qwen-3-0.6B.}
\label{tab:different_model_mae_results}
\resizebox{\textwidth}{!}{ 
\begin{tabular}{ll|l|ccc|ccc}
\toprule
\multirow{2}{*}{\textbf{Dataset}} & \multirow{2}{*}{\textbf{Event Type}} & \multirow{2}{*}{\textbf{Text Source}} & \multicolumn{3}{c|}{\textbf{RNN-based}} & \multicolumn{3}{c}{\textbf{Transformer-based}} \\
\cmidrule(lr){4-6} \cmidrule(lr){7-9}
& & & \textbf{\makecell{LLaMA-\\3.1-8B}} & \textbf{\makecell{LLaMA-\\3.2-1B}} & \textbf{\makecell{Qwen-\\3-0.6B}} & \textbf{\makecell{LLaMA-\\3.1-8B}} & \textbf{\makecell{LLaMA-\\3.2-1B}} & \textbf{\makecell{Qwen-\\3-0.6B}} \\
\midrule

\multirow{8}{*}{\textbf{Speech}} & \multirow{4}{*}{\textbf{New-word}} 
& Human & 1.726 & 1.718 & 1.729 & 1.605 & 1.594 & 1.611 \\
& & Claude-Sonnet-4.5 & 1.019 & 1.018 & 1.024 & 0.972 & 0.976 & 0.981 \\
& & GPT-4.1 & 1.009 & 1.005 & 1.011 & 0.962 & 0.966 & 0.966 \\
& & Qwen-Plus & 0.859 & 0.863 & 0.871 & 0.828 & 0.832 & 0.834 \\
\cmidrule(lr){2-9}

& \multirow{4}{*}{\textbf{\makecell{High-surprisal-\\token}}} 
& Human & 1.780 & 1.865 & 1.820 & 2.070 & 2.010 & 1.802 \\
& & Claude-Sonnet-4.5 & 1.624 & 1.554 & 1.530 & 1.605 & 1.541 & 1.536 \\
& & GPT-4.1 & 1.322 & 1.463 & 1.589 & 1.303 & 1.449 & 1.586 \\
& & Qwen-Plus & 2.003 & 1.838 & 1.968 & 1.945 & 1.831 & 1.988 \\
\midrule

\multirow{8}{*}{\textbf{\makecell{Writing \\ Prompt}}} & \multirow{4}{*}{\textbf{New-word}} 
& Human & 1.462 & 1.461 & 1.470 & 1.344 & 1.349 & 1.358 \\
& & Claude-Sonnet-4.5 & 1.265 & 1.277 & 1.279 & 1.167 & 1.173 & 1.183 \\
& & GPT-4.1 & 1.118 & 1.122 & 1.132 & 1.042 & 1.046 & 1.056\\
& & Qwen-Plus & 1.075 & 1.086 & 1.090 & 1.009 & 1.016 & 1.024 \\
\cmidrule(lr){2-9}

& \multirow{4}{*}{\textbf{\makecell{High-surprisal-\\token}}} 
& Human & 1.569 & 1.782 & 1.691 & 1.528 & 1.742 & 1.658 \\
& & Claude-Sonnet-4.5 & 2.285 & 2.025 & 1.901 & 2.183 & 1.958 & 1.863 \\
& & GPT-4.1 & 1.413 & 1.476 & 1.579 & 1.386 & 1.454 & 1.555 \\
& & Qwen-Plus & 1.799 & 1.613 & 1.805 & 1.746 & 1.572 & 1.778 \\

\bottomrule
\end{tabular}
}
\end{table*}

\subsection{Architectural Variations: Transformer-based Neural TPP}
\label{sec:tansformer-based}

% While the RNN-based architecture effectively captures local temporal dependencies within a fixed window, linguistic structures often exhibit long-range dependencies that span entire paragraphs.
% Inspired by recent self-attentive point processes \citep{zuo2020transformer, zhang2020self, DBLP:conf/nips/Panos24}, we introduce a Transformer-based neural TPP architecture that aggregates information across the entire event history to address this limitation. Specifically, our framework is a direct adaptation of the model proposed by \citet{zuo2020transformer}.

While RNN-based architectures effectively capture local temporal dependencies within a fixed window, they struggle with long-range dependencies, prompting the widespread adoption of Transformer-based models in recent works \citep{zuo2020transformer, zhang2020self, DBLP:conf/nips/Panos24}. 
Furthermore, linguistic structures often exhibit long-range dependencies that span entire paragraphs. Therefore, we primarily adapt the self-attentive neural TPP architecture from \citet{zuo2020transformer} to model natural language.

For a sequence of $L$ events, the initial input sequence is constructed from the absolute time step encodings $p_i$, contextual semantic embedding $e_i$, and the historical event interval encodings $v_i$, where $v_i$ is obtained by passing the raw intervals $x_i$ through a simple linear layer. 
Consequently, the initial hidden state for the sequence is formulated as $\tilde{\bm{H}}^{(0)} = \{W_v v_i + W_e e_i + W_p p_i\}_{i=1}^L$, which is then fed into a stack of $N$ Transformer encoder layers equipped with a causal masking mechanism to strictly prevent information leakage from future events.
Specifically, at layer $\ell$, the hidden state $\tilde{\bm{H}}^{(\ell)} $ is computed as:
\begin{equation}
    \tilde{\bm{H}}^{(\ell)} = \text{TransformerLayer}(\tilde{\bm{H}}^{(\ell-1)}).
\end{equation}
For the $i$-th event, the output hidden state vector $\tilde{\bm{h}}_i^{(N)}$ (the $i$-th element of $\tilde{\bm{H}}^{(N)}$) is then fed into the cumulative hazard function network to compute $\Phi(\tau|\tilde{\bm{h}}_i^{(N)})$.

As shown in \Cref{tab:different_model_mae_results}, the Transformer-based neural TPP achieves comparable, often slightly better MAE than its RNN counterpart, with minor exceptions (e.g., high-surprisal tokens in human-spoken texts).
Consistent with the RNN results, new-word events yield lower MAE with minimal variance across different embedding models, whereas the prediction of high-surprisal-token events remains more challenging and exhibits higher sensitivity to the choice of the LLM.

\begin{figure*}[!htbp]
\centering
\begin{subfigure}[b]{0.8\textwidth}
    \centering
    \includegraphics[width=\linewidth]{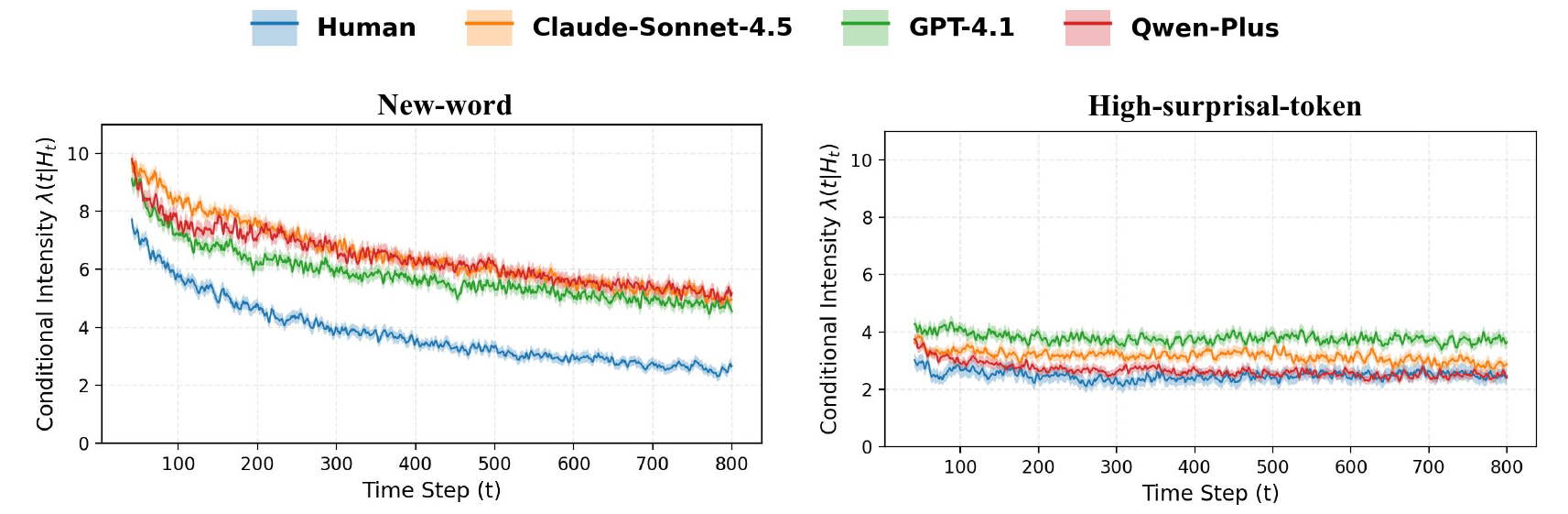}
    \caption{Speech dataset.}
    \label{fig:D-5_llama-8b_intensity_speech}
\end{subfigure}
\hfill
\begin{subfigure}{0.8\textwidth}
    \centering
    \includegraphics[width=\linewidth]{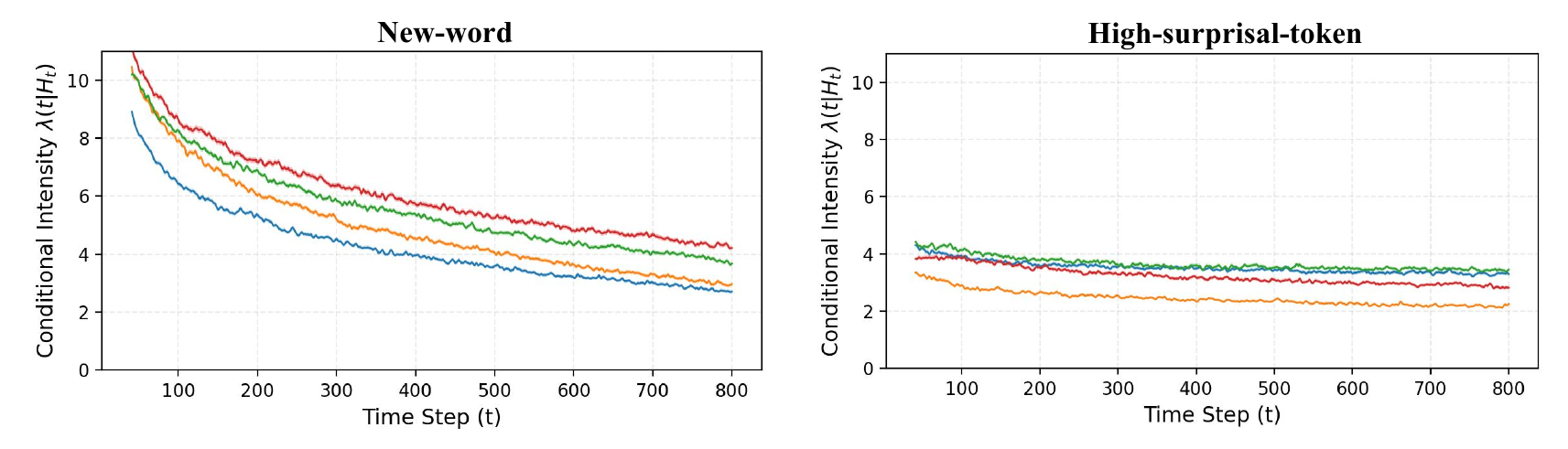}
    \caption{Writing-Prompts dataset.}
    \label{fig:D-5_llama-8b_intensity_writing}
\end{subfigure}
\caption{Estimated conditional intensity functions across different linguistic event types.
    All subplots utilize \textbf{Transformer-based} neural TPP incorporating semantic embeddings extracted via LLaMA-3.1-8B. Shadow areas are 95\% bootstrap confidence intervals.}
\label{fig:D-5_llama-8b_trans_intensity}
\end{figure*}

\begin{figure*}[!htbp]
\centering
\begin{subfigure}[b]{0.8\textwidth}
    \centering
    \includegraphics[width=\linewidth]{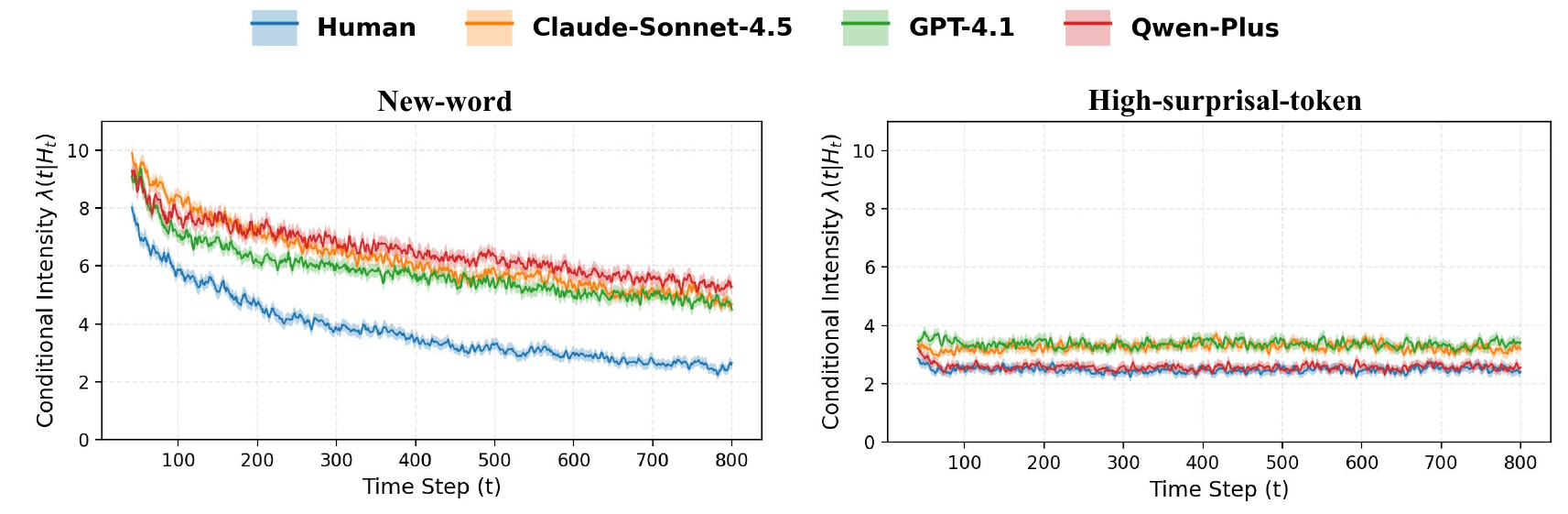}
    \caption{Speech dataset.}
    \label{fig:D-5_llama-1b_intensity_speech}
\end{subfigure}
\hfill
\begin{subfigure}{0.8\textwidth}
    \centering
    \includegraphics[width=\linewidth]{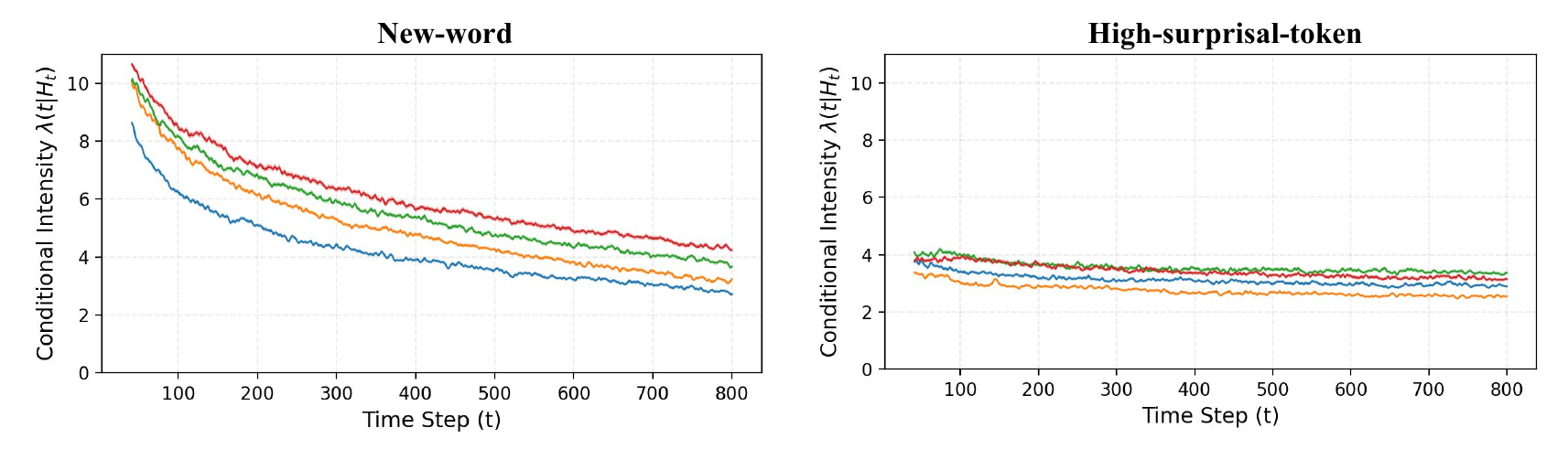}
    \caption{Writing-Prompts dataset.}
    \label{fig:D-5_llama-1b_intensity_writing}
\end{subfigure}
\caption{Estimated conditional intensity functions across different linguistic event types.
    All subplots utilize \textbf{Transformer-based} neural TPP incorporating semantic embeddings extracted via LLaMA-3.2-1B. Shadow areas are 95\% bootstrap confidence intervals.}
\label{fig:D-5_trans_llama-1b_intensity}
\end{figure*}

\begin{figure*}[!htbp]
\centering
\begin{subfigure}[b]{0.8\textwidth}
    \centering
    \includegraphics[width=\linewidth]{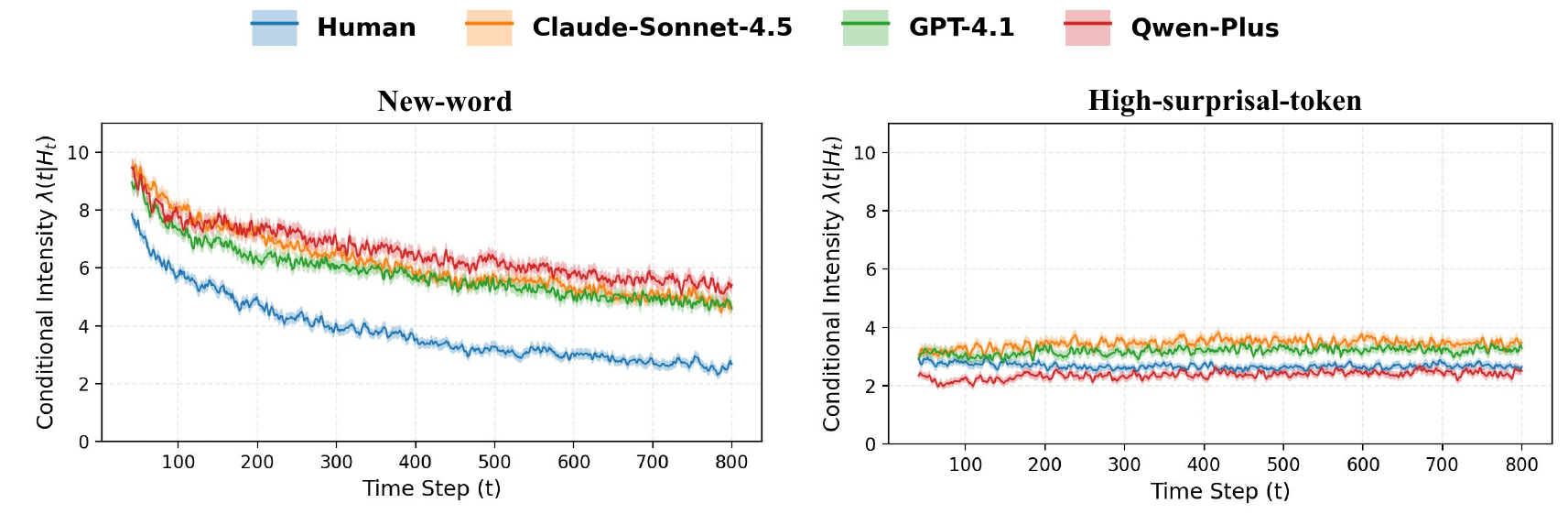}
    \caption{Speech dataset.}
    \label{fig:D-5_qwen-0.6b_intensity_speech}
\end{subfigure}
\hfill
\begin{subfigure}{0.8\textwidth}
    \centering
    \includegraphics[width=\linewidth]{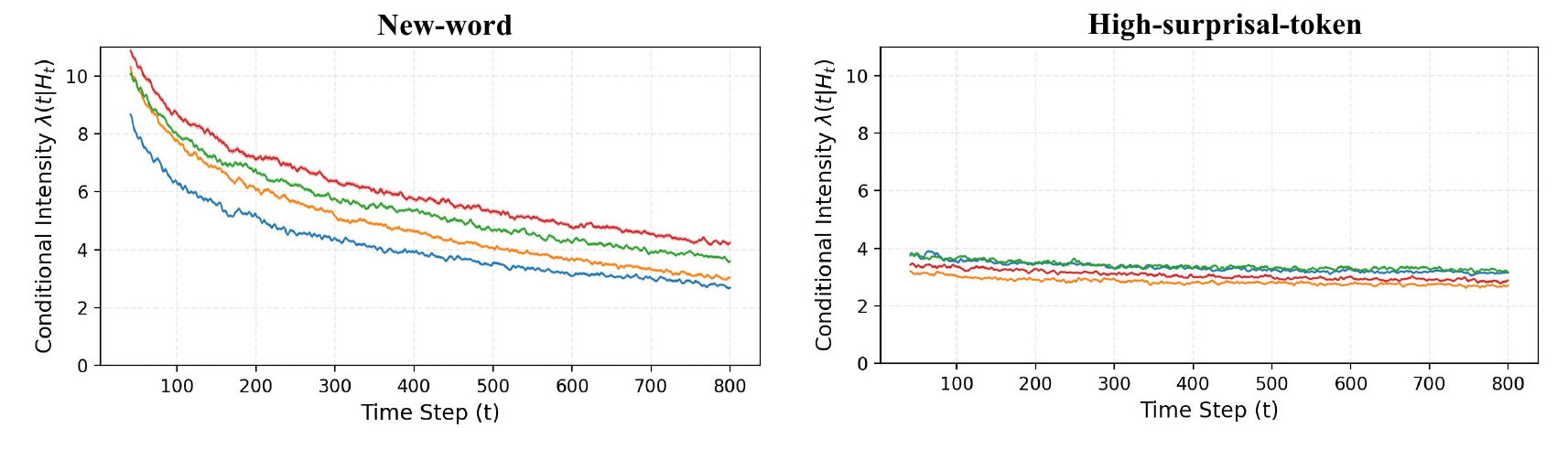}
    \caption{Writing-Prompts dataset.}
    \label{fig:D-5_qwen-0.6b_intensity_writing}
\end{subfigure}
\caption{Estimated conditional intensity functions across different linguistic event types.
    All subplots utilize \textbf{Transformer-based} neural TPP incorporating semantic embeddings extracted via Qwen-3-0.6B. Shadow areas are 95\% bootstrap confidence intervals.}
\label{fig:D-5_trans_qwen-0.6b_intensity}
\end{figure*}

Furthermore, \Cref{fig:D-5_llama-8b_trans_intensity} to \Cref{fig:D-5_trans_qwen-0.6b_intensity} show the conditional intensity curves lead to the same conclusion: the intensity of new-word events exhibits a continuous decay over time with LLMs persistently maintaining a higher generation rate than humans, while the intensity for high-surprisal tokens remains remarkably flat and stable.

\subsection{Full Classification Results}
\label{sec:appendix_all_cls_results}

\subsubsection{Conventional TPP Baseline}

To establish a classification baseline using the conventional TPP model, we randomly split each dataset into an 80\% training set and a 20\% testing set. The classification procedure relies on the statistical features of the fitted parameters and consists of the following stages:
\begin{enumerate}[leftmargin=10pt]
    \item \textbf{Feature Extraction and Statistical Modeling}: We fit the cumulative rate function $\lambda_C(t)$ to each training text to extract the five key parameters ($\lambda_1$ to $\lambda_5$). For each text source (Human vs. specific LLM), we compute the mean, standard deviation, and 95\% bootstrap confidence intervals (CIs) \citep{efron1992bootstrap, DBLP:books/sp/EfronT93} for these parameters.
    \item \textbf{Weighted Confidence Interval Scoring}: We assign a discriminative weight to each parameter based on the Intersection over Union (IoU) of its CIs between human and LLM texts ($Weight = 1 - IoU$). For each test text, we extract its parameters and compute matching scores for both classes. If a parameter falls within a class's CI, the class score increases by the feature weight. If it falls outside, the score is penalized by the absolute distance to the CI boundary divided by the class standard deviation (functioning as a Z-score penalty, capped at 3.0 to limit outlier impact). The text is assigned to the class with the higher comprehensive score.
\end{enumerate}

\begin{figure*}[!htbp]
\centering
\begin{subfigure}[b]{0.85\textwidth}
    \centering
    \includegraphics[width=\linewidth]{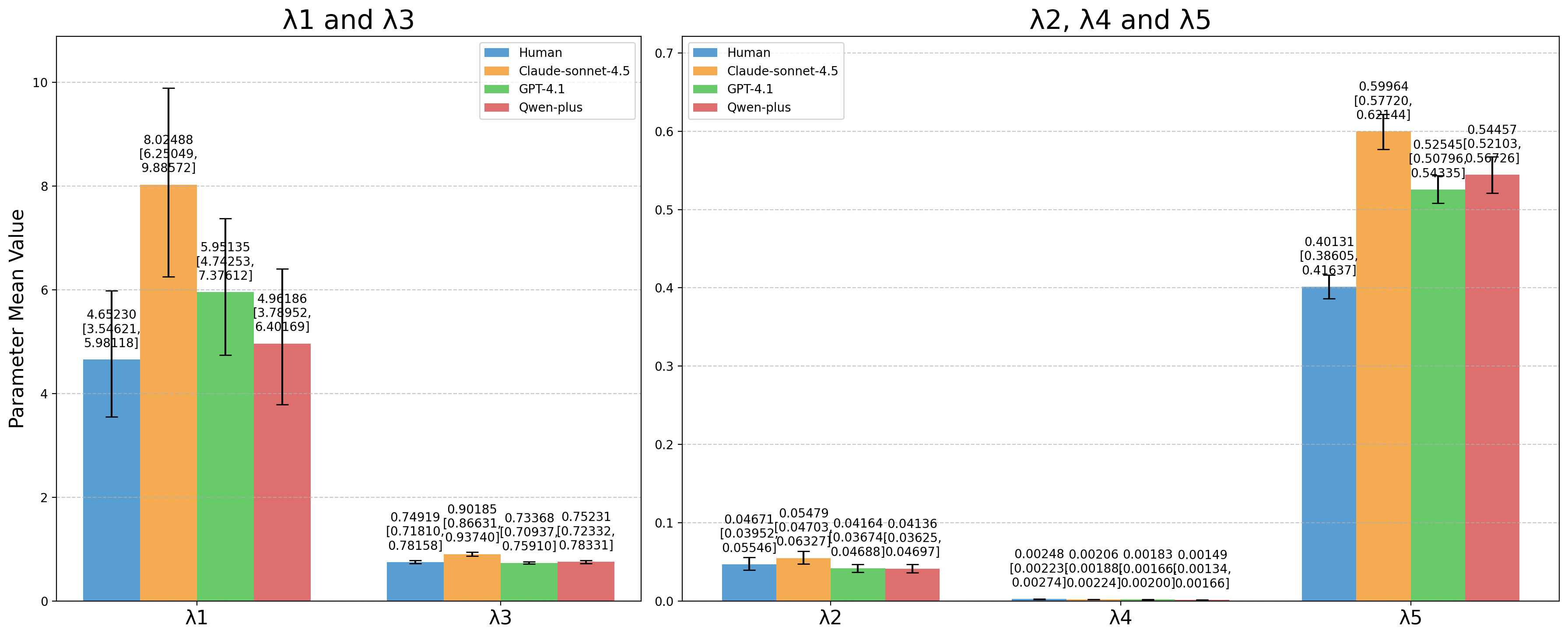}
    \caption{Speech dataset.}
\end{subfigure}
\hfill
\begin{subfigure}{0.85\textwidth}
    \centering
    \includegraphics[width=\linewidth]{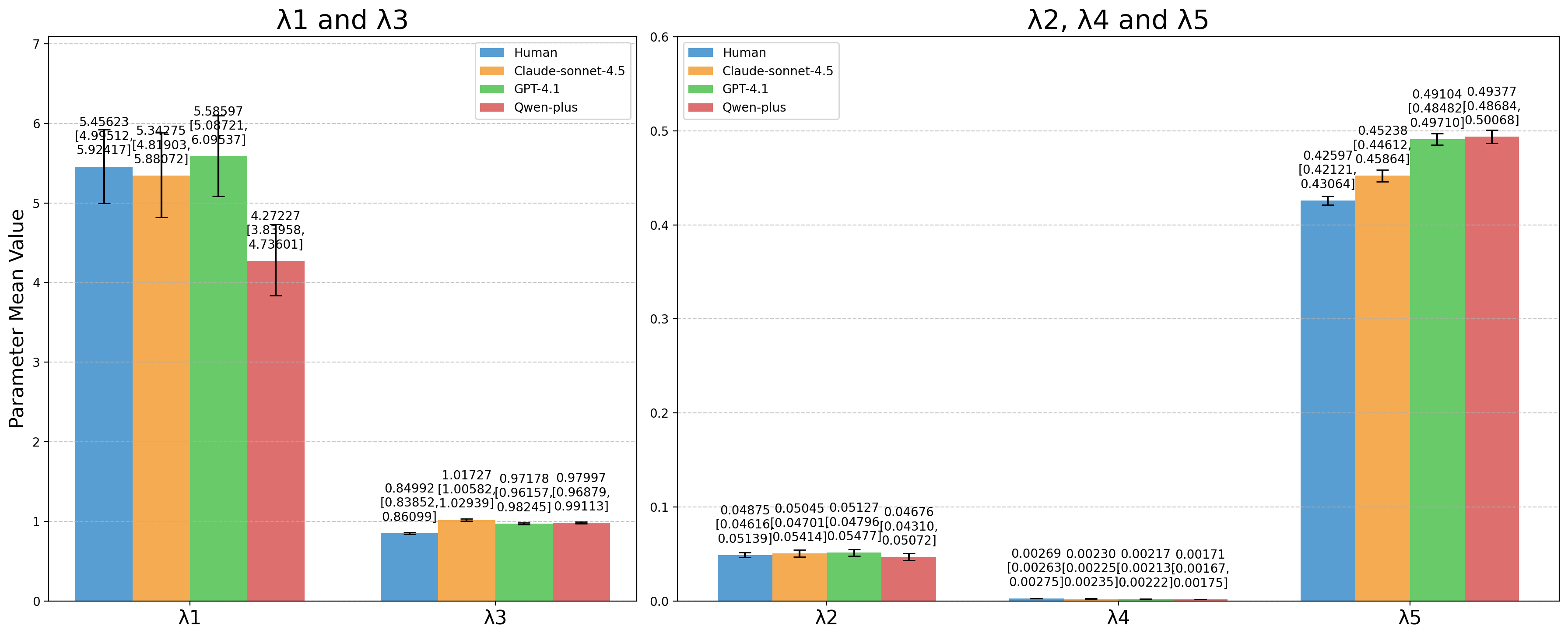}
    \caption{Writing-Prompts dataset.}
\end{subfigure}
\caption{Mean values and 95\% bootstrap confidence intervals of the five parameters ($\lambda_1$ to $\lambda_5$) estimated by the Conventional TPP model across different text sources on the Speech and Writing Prompt datasets.}
\label{fig:confidence_intervals_five}
\end{figure*}

The statistical distributions of the fitted parameters and the classification results are presented in Figure \ref{fig:confidence_intervals_five} and Table \ref{tab:cls_conventional_TPP_results}, respectively.

% \begin{figure*}[!htbp]
% % TODO: change color
%     \centering
%     \includegraphics[width=0.9\textwidth]{figs/hazard_plot/confidence_intervals_five.pdf}
%     \caption{Mean values and 95\% confidence intervals of the five parameters ($\lambda_1$ to $\lambda_5$) estimated by the Conventional TPP model across different text sources on the Speech and Writing Prompt datasets.}
%     \label{fig:confidence_intervals_five}
% \end{figure*}

\begin{table*}[!htbp]
\centering
\caption{Human vs. LLM classification results based on features extracted from the intensity functions estimated by the Conventional TPP model. We run separate binary classifications for each LLM (Claude-Sonnet-4.5, GPT-4.1, Qwen-Plus) against human texts.}
\label{tab:cls_conventional_TPP_results}
\resizebox{0.9\textwidth}{!}{%
\begin{tabular}{l|cc|ccc|ccc}
\toprule
\multirow{2}{*}{\textbf{Model}} & \multirow{2}{*}{\textbf{Accuracy}} & \multirow{2}{*}{\textbf{Macro F1}} & \multicolumn{3}{c|}{\textbf{Machine (Generated)}} & \multicolumn{3}{c}{\textbf{Human}} \\
\cmidrule(lr){4-6} \cmidrule(lr){7-9}
& & & Precision & Recall & F1 & Precision & Recall & F1 \\
\midrule
\multicolumn{9}{c}{\textbf{Speech Dataset}} \\
\midrule
Claude-sonnet-4.5 & 0.763 & 0.760 & 0.700 & 0.800 & 0.750 & 0.820 & 0.730 & 0.780 \\
GPT-4.1 & 0.725 & 0.720 & 0.570 & 0.820 & 0.680 & 0.880 & 0.670 & 0.760 \\
Qwen-plus & 0.725 & 0.720 & 0.620 & 0.780 & 0.690 & 0.820 & 0.690 & 0.750 \\
\midrule
\multicolumn{9}{c}{\textbf{Writing Prompt Dataset}} \\
\midrule
Claude-sonnet-4.5 & 0.656 & 0.650 & 0.740 & 0.630 & 0.680 & 0.570 & 0.690 & 0.630 \\
GPT-4.1 & 0.685 & 0.680 & 0.730 & 0.670 & 0.700 & 0.640 & 0.700 & 0.670 \\
Qwen-plus & 0.706 & 0.700 & 0.860 & 0.660 & 0.740 & 0.560 & 0.800 & 0.650 \\
\bottomrule
\end{tabular}
}
\end{table*}

\subsubsection{Our Neural TPP}
% The extracted features include the global statistics (e.g., mean, variance, skewness, kurtosis and percentiles), local statistics (e.g., mean and variance in different time bins) and temporal characteristics (e.g., number of peaks, volatility, and decay rate) of the estimated intensity functions. We train a Decision Tree classifier using these features to perform binary classification between human and LLM-generated texts for each LLM separately.
% Table \ref{tab:cls_rnn_results} presents the full classification results for the RNN-based neural TPP model with LLaMA-3.2-1B embeddings. The model achieves significantly higher accuracy and F1 scores compared to the Conventional TPP baseline, demonstrating the enhanced discriminative power of the neural TPP features.
Table \ref{tab:emb_comparison} presents the full classification results for the RNN-based neural TPP model across different embedding configurations. Across all the settings, we achieve consistently high accuracy and F1 scores, demonstrating the separability of the intensity curves estimated by the neural TPP models.

% \begin{table*}[htbp]
% \centering
% \caption{Human vs. LLM classification results based on features extracted from the intensity functions estimated by the RNN-based neural TPP with LLaMA-3.2-1B embeddings. We run separate binary classifications for each LLM (Claude-Sonnet-4.5, GPT-4.1, Qwen-Plus) against human texts.}
% \label{tab:cls_rnn_results}
% \resizebox{0.8\textwidth}{!}{%
% \begin{tabular}{l|cc|ccc|ccc}
% \toprule
% \multirow{2}{*}{\textbf{Model}} & \multirow{2}{*}{\textbf{Accuracy}} & \multirow{2}{*}{\textbf{Macro F1}} & \multicolumn{3}{c|}{\textbf{Machine (Generated)}} & \multicolumn{3}{c}{\textbf{Human}} \\
% \cmidrule(lr){4-6} \cmidrule(lr){7-9}
% & & & Precision & Recall & F1 & Precision & Recall & F1 \\
% \midrule
% \multicolumn{9}{c}{\textbf{Speech Dataset}} \\
% \midrule
% Claude-sonnet-4.5 & 0.920 & 0.920 & 0.929 & 0.910 & 0.919 & 0.912 & 0.930 & 0.921 \\
% GPT-4.1 & 0.935 & 0.935 & 0.939 & 0.930 & 0.935 & 0.931 & 0.940 & 0.935 \\
% Qwen-plus & 0.910 & 0.910 & 0.887 & 0.940 & 0.913 & 0.936 & 0.880 & 0.907 \\
% \midrule
% \multicolumn{9}{c}{\textbf{Writing Prompt Dataset}} \\
% \midrule
% Claude-sonnet-4.5 & 0.794 & 0.794 & 0.795 & 0.791 & 0.793 & 0.792 & 0.796 & 0.794 \\
% GPT-4.1 & 0.822 & 0.822 & 0.807 & 0.846 & 0.826 & 0.838 & 0.798 & 0.818 \\
% Qwen-plus & 0.942 & 0.942 & 0.933 & 0.951 & 0.942 & 0.950 & 0.932 & 0.941 \\
% \bottomrule
% \end{tabular}
% }
% \end{table*}

\begin{table*}[!htbp]
\centering
\caption{Classification performance (Accuracy and AUROC) of the RNN-based neural TPP model across four embedding configurations (without-LLM, LLaMA-3.1-8B, LLaMA-3.2-1B, Qwen-3-0.6B) for each LLM vs. Human binary classification.}
\label{tab:emb_comparison}
\resizebox{0.9\textwidth}{!}{%
\begin{tabular}{l|cc|cc|cc|cc}
\toprule
\multirow{2}{*}{\textbf{Model}} & \multicolumn{2}{c}{\textbf{w/o LLM}} & \multicolumn{2}{c}{\textbf{LLaMA-3.1-8B}} & \multicolumn{2}{c}{\textbf{LLaMA-3.2-1B}} & \multicolumn{2}{c}{\textbf{Qwen-3-0.6B}} \\
\cmidrule(lr){2-3} \cmidrule(lr){4-5} \cmidrule(lr){6-7} \cmidrule(lr){8-9}
& \textbf{Acc} & \textbf{AUROC} & \textbf{Acc} & \textbf{AUROC} & \textbf{Acc} & \textbf{AUROC} & \textbf{Acc} & \textbf{AUROC} \\
\midrule
\multicolumn{9}{c}{\textbf{Speech Dataset}} \\
\midrule
Claude-sonnet-4.5 & 0.9750 & 0.9964 & 0.9450 & 0.9913 & 0.9500 & 0.9896 & 0.9600 & 0.9914 \\
GPT-4.1 & 0.9800 & 0.9994 & 0.9650 & 0.9932 & 0.9600 & 0.9939 & 0.9750 & 0.9967 \\
Qwen-plus & 0.9950 & 0.9999 & 0.9800 & 0.9988 & 0.9800 & 0.9978 & 0.9800 & 0.9990 \\
\midrule
\multicolumn{9}{c}{\textbf{Writing Prompt Dataset}} \\
\midrule
Claude-sonnet-4.5 & 0.8615 & 0.9332 & 0.8535 & 0.9263 & 0.8735 & 0.9464 & 0.8080 & 0.8937 \\
GPT-4.1 & 0.9225 & 0.9734 & 0.9140 & 0.9719 & 0.8900 & 0.9582 & 0.8915 & 0.9610 \\
Qwen-plus & 0.9745 & 0.9958 & 0.9585 & 0.9937 & 0.9705 & 0.9952 & 0.9525 & 0.9913 \\
\bottomrule
\end{tabular}
}
\end{table*}

\subsubsection{Feature Importance Analysis}
\label{sec:appendix_feature_importance}
% TODO: fix the discussion
% The feature importance heatmap in \Cref{fig:feature_importance_heatmap} includes the different combinations of embedding types (LLaMA-3.2-1B, Qwen-3-0.6B and no-embedding) and generation models (Claude-Sonnet-4.5, GPT-4.1, Qwen-Plus) on both datasets. Overall, the temporal characteristics tend to be the most discriminative features. Number of peaks and volatility are particularly important, suggesting that the rhythmicity and stability of new-word introduction are key factors distinguishing human from LLM-generated texts. For global statistics, low percentiles show higher importance than other features. For local statistics, the mean and variance in the early time bins (0-20\% of the text) are more important than those in later bins, which may reflect that early-stage temporal dynamics are more distinctive between human and LLM generation.

\begin{figure*}[!htbp]
    \centering
    \includegraphics[width=1\textwidth]{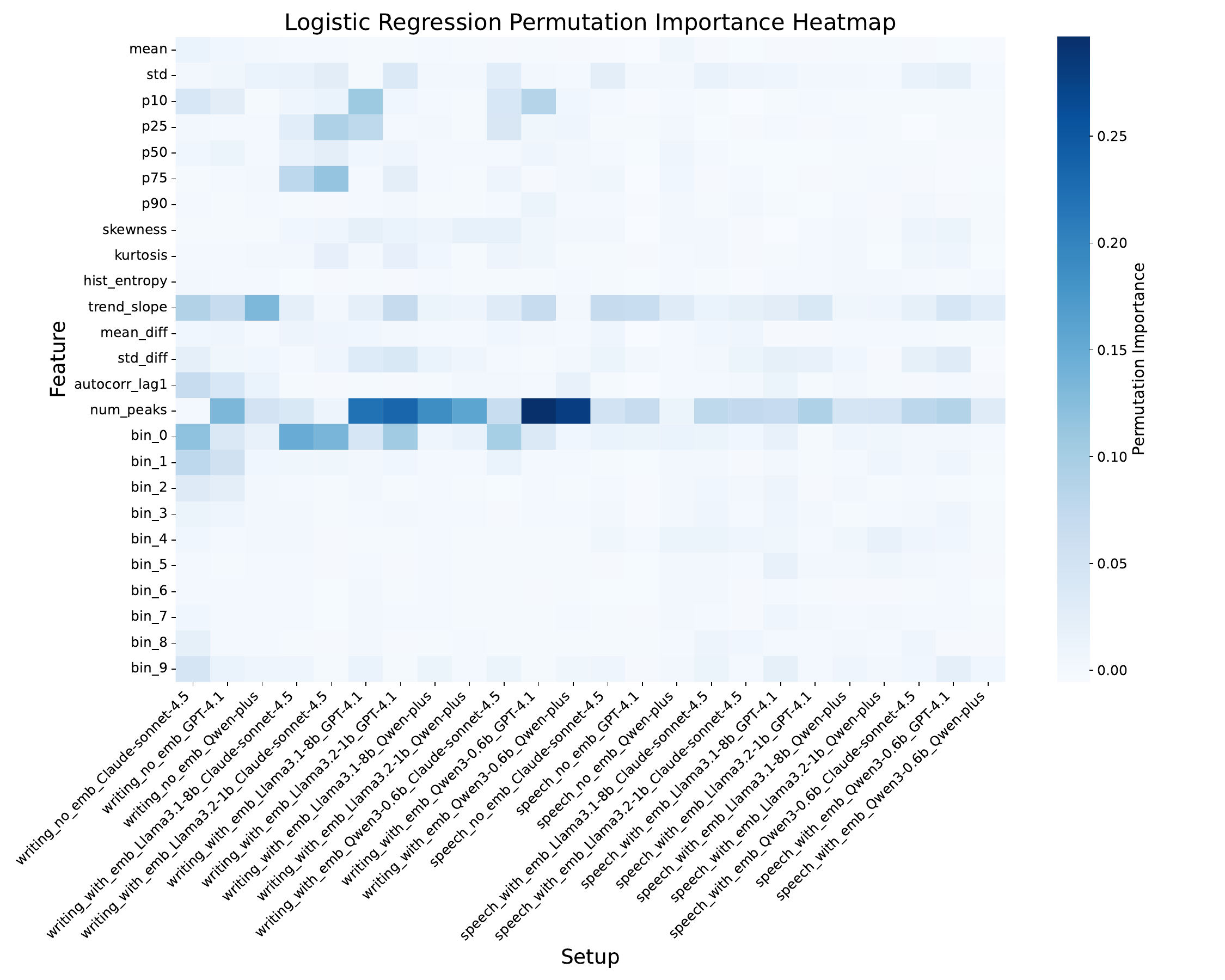}
    \caption{Feature importance heatmap for the Logistic Regression classifiers applied to the RNN-based neural TPP across all combinations of embedding types (without-LLM and LLaMA-3.1-8B, LLaMA-3.2-1B, Qwen-3-0.6B) and generation models (Claude-Sonnet-4.5, GPT-4.1, Qwen-Plus) on both datasets. The importance scores are estimated using the permutation importance method. The color intensity represents the importance of each feature in distinguishing human from LLM-generated texts.}
    \label{fig:feature_importance_heatmap}
\end{figure*}

The feature importance heatmap in \Cref{fig:feature_importance_heatmap} includes the different combinations of embedding types (without-LLM, LLaMA-3.1-8B, LLaMA-3.2-1B and Qwen-3-0.6B) and generation models (Claude-Sonnet-4.5, GPT-4.1, Qwen-Plus) on both datasets. We use the permutation importance method to estimate the importance scores of each feature.

% % include heatmap of feature importance for all setups
% \begin{figure*}[!htbp]
%     \centering
%     \includegraphics[width=1\textwidth]{figs/cls/feature_importance_heatmap.pdf}
%     \caption{Feature importance heatmap for the Decision Tree classifiers across all combinations of embedding types (LLaMA-3.2-1B, Qwen-3-0.6B and no-embedding) and generation models (Claude-Sonnet-4.5, GPT-4.1, Qwen-Plus) on both datasets. The features are categorized into global statistics, local statistics, and temporal characteristics. The color intensity represents the importance of each feature in distinguishing human from LLM-generated texts.}
%     \label{fig:feature_importance_heatmap}
% \end{figure*}

% \begin{figure*}[!htbp]
%     \centering
%     \includegraphics[width=1\textwidth]{figs/cls/top3_features_tree_scatter_speech_llama_qwen.pdf}
%     \caption{The classification tree and corresponding scatter plots of the top-3 most important features for the binary classification of human vs. Qwen-plus texts on the Speech dataset, using the Transformer-based neural TPP with LLaMA-3.2-1B embeddings.}
%     \label{fig:cls_scatter_plots_speech_llama_qwen}
% \end{figure*}

\subsection{Visualization of New Word Occurrences}

\begin{figure*}[!htbp]
    \centering
    \includegraphics[width=1\textwidth]{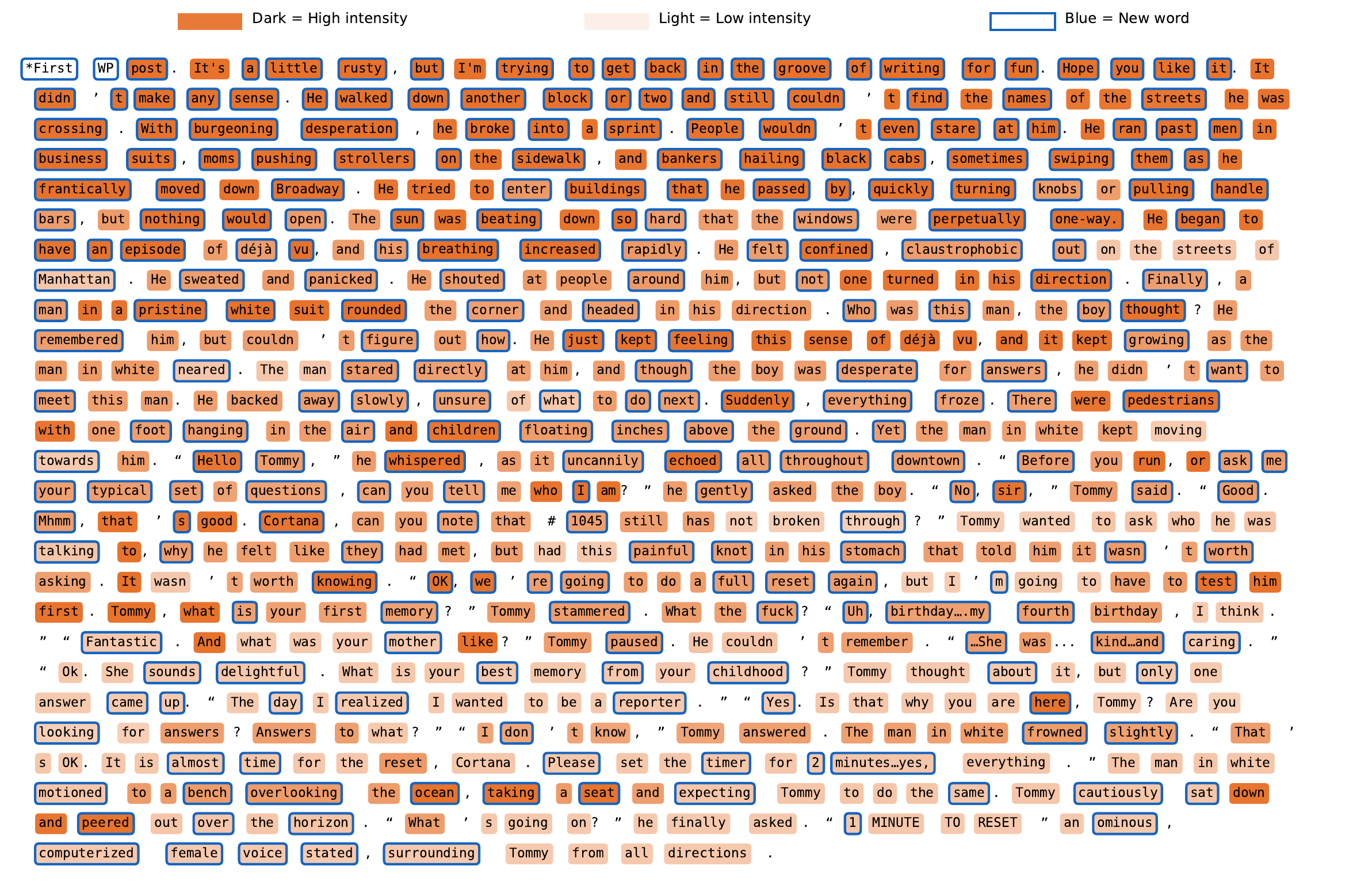}
    \caption{Heatmap of new-word occurrences in a \texttt{Writing-Prompts} sample (Example 1). Darker orange gradients indicate higher intensity of each word estimated by the RNN-based neural TPP model, whereas new words (appearing for the first time in the context) are denoted by blue bounding boxes. The visualizations demonstrate the bursty, non-uniform distribution of new-word events across the generation sequence.}
    % with clusters corresponding to topic shifts and narrative development.}
    \label{fig:heatmap_comparison_1}
\end{figure*}

\begin{figure*}[!htbp]
    \centering
    \includegraphics[width=1\textwidth]{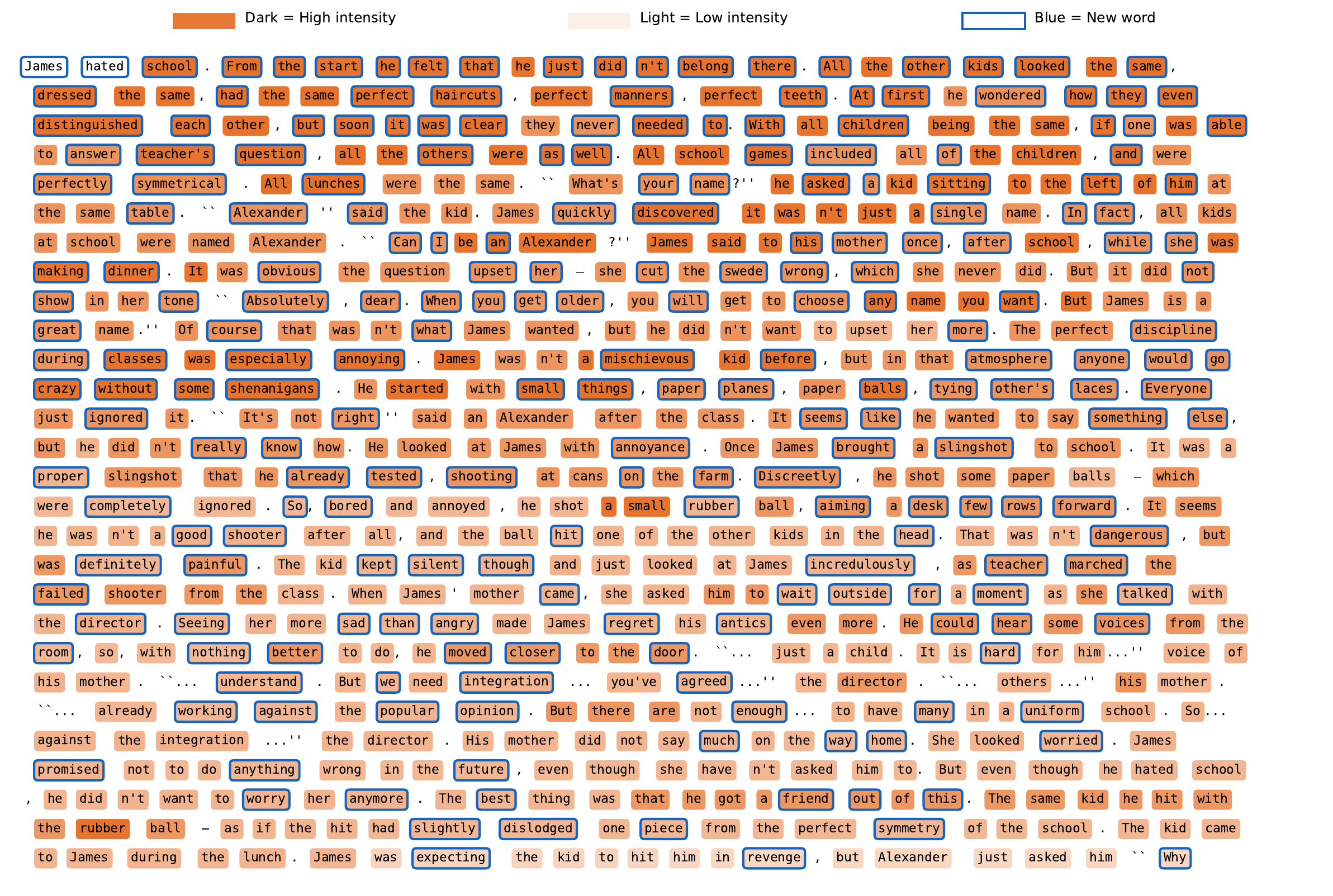}
    \caption{Heatmap visualizations of new-word occurrence patterns in a second human-written \texttt{Writing-Prompts} sample (Example 2), with intensity levels estimated by the RNN-based neural TPP model.}
    \label{fig:heatmap_comparison_2}
\end{figure*}

To provide intuitive insight into how new-word events are distributed during text generation, we present heatmap visualizations of representative samples from \texttt{Writing-Prompts} dataset ( \Cref{fig:heatmap_comparison_1} and \Cref{fig:heatmap_comparison_2}), 
which directly illustrate the temporal patterns quantified by RNN-based neural TPP models.

In each heatmap, words appearing for the first time in the context are highlighted, allowing visual inspection of new word introduction patterns. The visualizations demonstrate how new word events cluster and disperse across different discourse structures. New words tend to appear in bursts corresponding to topic introductions and narrative developments, rather than being uniformly distributed. This bursty, non-stationary behavior validates our use of sophisticated neural TPP models over simple Poisson assumptions.

The heatmaps also reveal the heterogeneity in temporal patterns across different texts, even within the same human-written genre. Some texts exhibit rapid early saturation with sparse later novelty, while others maintain more sustained lexical innovation. This variability underscores the importance of modeling conditional intensity functions that adapt to contextual and semantic information, as implemented in our RNN-based architectures.

% \begin{figure*}[!htbp]
%     \centering
%     \begin{minipage}{0.48\textwidth}
%         \centering
%         \includegraphics[width=\linewidth]{figs/demo/20.pdf}
%     \end{minipage}
%     \hfill
%     \begin{minipage}{0.48\textwidth}
%         \centering
%         \includegraphics[width=\linewidth]{figs/demo/38.pdf}
%     \end{minipage}
%     \caption{Heatmap visualizations of new word occurrence patterns in speech samples. New words (appearing for the first time in the context) are highlighted in color. The visualizations demonstrate the bursty, non-uniform distribution of new word events across the generation sequence, with clusters corresponding to topic shifts and narrative development.}
%     \label{fig:heatmap_comparison}
% \end{figure*}

% \begin{figure*}[!htbp]
%     \centering
%     \includegraphics[width=1\textwidth]{figs/cls/top3_features_tree_scatter_writing_llama_qwen.pdf}
%     \caption{The classification tree and corresponding scatter plots of the top-3 most important features for the binary classification of human vs. Qwen-plus texts on the Writing Prompt dataset, using the Transformer-based neural TPP with LLaMA-3.2-1B embeddings.}
%     \label{fig:cls_scatter_plots_writing_llama_qwen}
% \end{figure*}

\end{document}